\documentclass[journal,twoside,web]{ieeecolor}
\usepackage{generic}
\usepackage{cite}
\usepackage{graphicx}
\usepackage{textcomp}

\usepackage[utf8]{inputenc}         
\usepackage[english]{babel}
\usepackage{multirow}

\usepackage{multicol}

\usepackage{soul}

\usepackage[normalem]{ulem}

\usepackage{epstopdf}
\graphicspath{{./image/}}

\usepackage[cmex10]{amsmath}
\usepackage{amssymb}

\usepackage{array}
\usepackage{rotating}

\DeclareMathOperator*{\argmin}{argmin}

\usepackage{makecell}
\usepackage{subcaption}
\usepackage{textcomp}

\usepackage{lipsum}
\usepackage{url}

\usepackage{cite}
\usepackage{capt-of}

\usepackage{tabularx}
\usepackage{bm}
\usepackage{calc}

\usepackage{xcolor,colortbl}

\usepackage{tabu}
\usepackage{mdframed}

\usepackage{pifont}

\def\BibTeX{{\rm B\kern-.05em{\sc i\kern-.025em b}\kern-.08em
    T\kern-.1667em\lower.7ex\hbox{E}\kern-.125emX}}
\begin{document}
\title{Towards More Accurate Automatic Sleep Staging via Deep Transfer Learning}
\author{{Huy~Phan$^{*}$, Oliver~Y.~Ch\'{e}n, Philipp Koch, Zongqing Lu, Ian McLoughlin, Alfred Mertins, and~Maarten~De~Vos
\thanks{H. Phan is with the School of Electronic Engineering and Computer Science, Queen Mary University of London, London E1 4NS, UK. I. McLouglin is with Singapore Institute of Technology, Singapore 138683. O. Y. Ch\'{e}n is with the Institute of Biomedical Engineering, University of Oxford, Oxford OX3 7DQ, UK. Z. Lu is with the Department of Computer Science, Peking University, Beijing, 100080, China. P. Koch and A. Mertins are with the Institute for Signal Processing, University of L\"ubeck, L\"ubeck 23562, Germany. M. De Vos is with the Department of Electrical Engineering and the Department of Development and Regeneration, KU Leuven, 3001 Leuven, Belgium.}
\thanks{$^*$Corresponding author: {\tt\footnotesize h.phan@qmul.ac.uk}}
}}

\maketitle

\begin{abstract}
\emph{Background:} Despite recent significant progress in the development of automatic sleep staging methods, building a good model still remains a big challenge for sleep studies with a small cohort due to the \emph{data-variability} and \emph{data-inefficiency} issues. This work presents a deep transfer learning approach to overcome these issues and enable transferring knowledge from a large dataset to a small cohort for automatic sleep staging. \emph{Methods:} We start from a generic end-to-end deep learning framework for sequence-to-sequence sleep staging and derive two networks as the means for transfer learning. The networks are first trained in the source domain (i.e. the large database). The pretrained networks are then finetuned in the target domain (i.e. the small cohort) to complete knowledge transfer. We employ the Montreal Archive of Sleep Studies (MASS) database consisting of 200 subjects as the source domain and study deep transfer learning on three different target domains: the Sleep Cassette subset and the Sleep Telemetry subset of the Sleep-EDF Expanded database, and the Surrey-cEEGrid database. The target domains are purposely adopted to cover different degrees of data mismatch to the source domains. \emph{Results:} Our experimental results show significant performance improvement on automatic sleep staging on the target domains achieved with the proposed deep transfer learning approach. \emph{Conclusions:} These results suggest the efficacy of the proposed approach in addressing the above-mentioned data-variability and data-inefficiency issues. \emph{Significance:} As a consequence, it would enable one to improve the quality of automatic sleep staging models when the amount of data is relatively small.
\footnote{\footnotesize The source code and the pretrained models are published at \url{http://github.com/pquochuy/sleep_transfer_learning}.}
\end{abstract}

\begin{IEEEkeywords}
Automatic sleep staging, sequence-to-sequence, deep learning, transfer learning. 
\end{IEEEkeywords}

\section{Introduction}
\label{sec:introduction}

Sleep scoring \cite{Iber2007,Hobson1969} aims to determine  sleep stages from polysommography (PSG) recordings. In clinical environments, this task has been mainly performed manually by clinicians following developed guidelines \cite{Iber2007,Hobson1969}. Since the manual scoring is time-consuming, costly, and prone to human errors, automating the scoring process has been a long-lasting focus in the sleep research community \cite{Redmond2006,Phan2019a,Phan2019b,Supratak2017,Tsinalis2016, Stephansen2018,Mikkelsen2018,Chambon2018}. Automatic sleep scoring is particularly important in home-based sleep monitoring \cite{Mikkelsen2019,Looney2016,Goverdovsky2016,Kidmose2013}. Recent years have seen a new generation of mobile electroencephalography (EEG) devices that provide a cost-effective solution to screen a wide population for epidemiological studies and to monitor specific populations at risk of sleep disorders.

Deep learning has been successfully applied to numerous domains and has received much attention from the sleep research community. Past work has looked at various deep network architectures, such as deep neural networks (DNNs) \cite{Dong2017}, convolutional neural networks (CNNs) \cite{Tsinalis2016, Andreotti2018b, phan2018c, Chambon2018, Vilamala2017, Sors2018, Stephansen2018, Patanaik2018, Biswal2018a, Biswal2017, Sun2017, Olesen2018}, and recurrent neural networks (RNNs) \cite{phan2018d,Koch2018a,Koch2018b, Zhai2020}, and novel ways to carry out sleep staging like sequence-to-sequence classification scheme \cite{Phan2019a, Supratak2017}. Reviews of the most recent progress on deep learning for automatic sleep staging can be found in \cite{Faust2019, Roy2019, Fiorillo2019}. However, considerably less attention has been paid to make sleep staging models more robust to the challenges of sleep data variability and to make them data-efficient (i.e. using less data). Despite the fact that the performance of machine's sleep staging has been on par with manual scoring by sleep experts \cite{Phan2019a, Stephansen2018, Supratak2017, Patanaik2018, Biswal2018a, Biswal2017, Sun2017}, we have not seen it widely adopted clinically. This is arguably due to two major technical drawbacks of the sleep staging models: \emph{data variability} and \emph{data inefficiency}.

\setlength\tabcolsep{1pt}
\begin{table}[!b]
	\vspace{-0.25cm}
	\caption{Out-domain performance of the single-channel SeqSleepNet+ trained on MASS database in comparison to its in-domain performance.}
	\footnotesize
	\vspace{-0.2cm}
	\begin{center}
		\begin{tabular}{|>{\arraybackslash}m{0.45in}|>{\centering\arraybackslash}m{0.55in}|>{\centering\arraybackslash}m{0.72in}|>{\centering\arraybackslash}m{0.72in}|>{\centering\arraybackslash}m{0.72in}|>{\centering\arraybackslash}m{0in} @{}m{0pt}@{}}
			\cline{1-5}
			Database & MASS & Sleep-EDF-SC & Sleep-EDF-ST & Surrey-cEEGrid & \parbox{0pt}{\rule{0pt}{1ex+\baselineskip}} \\ [0ex]  	
			
			\cline{1-5}
			Input & C4-A1 & Fpz-Cz & Fpz-Cz & cEEGrid  & \parbox{0pt}{\rule{0pt}{0.25ex+\baselineskip}} \\ [0ex]  	
			\cline{1-5}
			Accuracy & \makecell{$84.5$ \\ (in-domain)} & \makecell{$81.2$ \\ (out-of-domain)} & \makecell{$80.5$ \\ (out-of-domain)} & \makecell{$10.6$ \\ (out-of-domain)}  & \parbox{0pt}{\rule{0pt}{0.25ex+\baselineskip}} \\ [0ex]  	
			\cline{1-5}
			Mismatch & - & slight & slight & severe  & \parbox{0pt}{\rule{0pt}{0.25ex+\baselineskip}} \\ [0ex]  	
			\cline{1-5}
		\end{tabular}
	\end{center}
	\label{tab:performance_indomain_outdomain}
\end{table}

\noindent{\bf Data variability:} PSG signals recorded in a particular recording setup are characterized by a number of parameters such as sensors' frequency response and output level, and signal processing applied to the raw signal. These factors contribute to the \emph{transfer function} of the recording device and affect how the physiological signals are converted into digital PSG output. As a result, sleep data recorded in different setups may have different transfer functions due to the variations in their underlying hardware and software processing pipelines. Furthermore, discrepancies in channel layouts \cite{Stephansen2018} and cohort characteristics \cite{Cooray2019} are also likely in different sleep studies. From the viewpoint of machine learning models, these variations and discrepancies lead to \emph{domain shift} or mismatch between sleep data sources. Data mismatch across different acquisition conditions are computationally significant, degrading the accuracy of sleep staging models on unseen data with a novel recording condition. Therefore, if a sleep staging model is deployed on an unseen sleep data whose properties differ from the data used for training the model, the data mismatch can result in poor inference performance. As evidenced in our experiments (cf. Section \ref{ss:experimental_results}) and shown in Table \ref{tab:performance_indomain_outdomain}, the single-channel SeqSleepNet \cite{Phan2019a} model trained on the MASS database suffers from an accuracy drop when it is evaluated \emph{out of domain} (i.e. being tested on other three databases Sleep-EDF-SC, Sleep-EDF-ST, and Surrey-cEEGrid) relative to the one obtained with \emph{in-domain} evaluation (i.e. via cross-validation on the MASS database itself). The performance loss depends on the level of data mismatch.

\noindent{\bf Data inefficiency:} Existing deep-learning based sleep staging models cannot escape from the curse of data inefficiency of the deep learning paradigm. That is, training a deep neural network generally requires a large amount of data. In fact, expert-level performance on automatic sleep staging is only obtainable with these models when the training cohort is large, i.e. hundreds or thousands of subjects \cite{Phan2019a, Stephansen2018}. The networks' performances decline significantly when they are trained with a small cohort (e.g. ten or twenty subjects \cite{Phan2019c, Supratak2017}). Unfortunately, in practice, many sleep studies only have access to a small cohort, in the order of a few dozens of subjects \cite{Kemp2000, Goldberger2000, Olesen2016, Cooray2019,Andreotti2018, Mikkelsen2019}. Thus, the small data in these studies hinder deep learning models to perform well. 

An easy and obvious solution for the above-mentioned obstacles is to collect training data from all types of recording setups (e.g. recording devices, channel layouts, and preprocessing softwares) that will be foreseeably encountered in the deployment phase. However, this is an expensive, time-consuming, and even infeasible solution. First, most of large sleep databases are proprietary, making those inaccessible for research purposes. Second, even if they are available, a huge effort would be required to score these data manually. Third, novel setups will likely emerge when one studies a particular sleep disorder \cite{Cooray2019,Andreotti2018} or when one explores the feasibility of a new monitoring device \cite{Mikkelsen2019}.

In this work, we present a practical solution based on transfer learning to tackle these obstacles, to build more accurate sleep staging models when the available data is small, and to recover the performance of the models otherwise lost due to data variability. We leverage a reasonably large sleep database, which is publicly available, and use a sleep-staging deep neural network as a device to transfer knowledge from this database to improve sleep staging performance on another small cohort with a different recording setup. More specifically, the network is firstly trained with the large database (the source domain) and subsequently finetuned with the small cohort (the target domain) to complete transfer learning. In this context, \emph{finetuning} means a part or the entire of the pretrained network is further trained with the target domain data. The main contributions of this work include:
\begin{itemize}
	\item A new perspective of looking at data variability and data inefficiency in the automatic sleep staging problem, and developing a deep transfer learning approach to overcome sleep data mismatch and enable knowledge transfer to improve sleep-staging performance on small cohorts. In-depth investigation into the influence of the number of subjects on the transfer learning performance was also conducted.
	\item The generalization of a sequence-to-sequence sleep staging framework from which two state-of-the-art models \emph{SeqSleepNet+} and \emph{DeepSleepNet+} are developed and used in the study.
	\item A systematic study highlighting different target domains with varying data-mismatch degrees to the source domain, different transfer learning scenarios (i.e. single-channel and multi-channel input), different finetuning strategies, and different state-of-the-art sleep staging models. Our transfer learning approach outperforms all the tested baselines and existing works in solving the automatic sleep staging  on the target sleep databases.
\end{itemize}

This work extends our preliminary work in \cite{Phan2019c} in several aspects. First, we study transfer learning with a wider spectrum of channel combinations for the networks' input rather than a single channel. Second, the studies in \cite{Phan2019c} employed SeqSleepNet \cite{Phan2019a} as the transfer learning device, here the studies are carried out on two different networks inherited from SeqSleepNet \cite{Phan2019a} and DeepSleepNet \cite{Supratak2017}. These two state-of-the-art networks are diverging in their architectures \cite{Phan2019d}; therefore, it is important to examine if these dissimilarities give rise to any difference in their performance and to explain their behaviors in transfer learning. Third, the work in \cite{Phan2019c} only studied deep transfer learning on the Sleep-EDF-SC as the target domain. Here, we cover multiple target domains with varying degrees of channel mismatch. Fourth, we study in-depth the influence of the number of target subjects on the transfer learning's performance.

\vspace{-0.15cm}
\section{Materials}
\vspace{-0.15cm}
\subsection{Source Domain}
We adopted the public Montreal Archive of Sleep Studies (MASS) database \cite{Oreilly2014} as the source domain in this study as it is sufficiently large. 

{\bf MASS:} This database was pooled from different hospital-based sleep laboratories, consisting of whole-night recordings from 200 subjects (97 males and 103 females) aged between 18 and 76 years. Manual annotation was accomplished by sleep experts according to the AASM standard \cite{Iber2007} (SS1 and SS3 subsets) or the R\&K standard \cite{Hobson1969}  (SS2, SS4, and SS5 subsets). As in \cite{Phan2019b,Phan2019a}, we converted different annotations into five sleep stages \{W, N1, N2, N3, and REM\} and expanded 20-second epochs into 30-second ones by including 5-second segments before and after each epoch. We used the C4-A1 EEG, ROC-LOC EOG, and CHIN1-CHIN2 EMG in our experiments.

\vspace{-0.15cm}
\subsection{Target Domains}

Three different sleep databases are used as the target domains. These adopted cohorts have diverging health conditions, i.e. healthy (Sleep-EDF-SC) vs. mild sleep difficulty (Sleep-EDF-ST) \cite{Kemp2000, Goldberger2000}, and channel characteristics (i.e. traditional PSG recording (Sleep-EDF-SC and Sleep-EDF-ST) vs. wearable around-the-ear EEG recordings (Surrey-cEEGrid) \cite{Mikkelsen2019, Sterr2018}).

{\bf Sleep-EDF-SC:} This is the Sleep Cassette (SC) subset of the Sleep-EDF Expanded dataset \cite{Kemp2000, Goldberger2000}, consisting of 20 subjects aged 25-34. Two subsequent day-night PSG recordings were collected for each subject, except for subject 13 who has only one-night data. Each 30-second PSG epoch was manually labelled into one of eight categories \{W, N1, N2, N3, N4, REM, MOVEMENT, UNKNOWN\} by sleep experts according to the R\&K standard \cite{Hobson1969}. Similar to previous works \cite{Tsinalis2016, Tsinalis2016b, Supratak2017, Phan2019b, phan2018c, phan2018d}, N3 and N4 stages were merged into a single stage N3 and MOVEMENT and UNKNOWN categories were excluded. Since full EMG recordings are not available, we only adopted the Fpz-Cz EEG and ROC-LOC EOG (i.e. the EOG horizontal) channels in this study. As this database has been used differently in literature, it should be stressed that only the in-bed parts (from \emph{lights off} time to \emph{lights on} time) of the recordings were used as recommended in \cite{Imtiaz2014,Imtiaz2015,Tsinalis2016, Tsinalis2016b, Phan2019b, phan2018c, phan2018d}.

{\bf Sleep-EDF-ST:} This is the Sleep Telemetry (ST) subset of the Sleep-EDF Expanded dataset\cite{Kemp2000, Goldberger2000} which was collected for studying the temazepam effects on sleep. The subset consists of 22 Caucasian subjects (7 males and 15 females) aged 18-79 with mild difficulty falling asleep. Although the PSG signals were recorded for two nights, one after temazepam intake and one after placebo intake, only the placebo nights are available. Manual annotation was done similar to the Sleep-EDF-SC subset. Beside Fpz-Cz EEG and ROC-LOC EOG, the submental EMG channel is available and additionally adopted. Similar to the the Sleep-EDF-SC subset, only the in-bed parts of the recordings were used.

\begin{figure} [!t]
	\centering
	\includegraphics[width=0.8\linewidth]{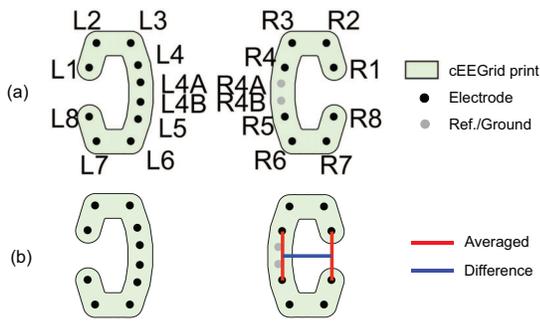}
	\vspace{-0.1cm}
	\caption{Illustration of (a) the cEEGrid electrode array which was used to record the Surrey-cEEGrid database \cite{Mikkelsen2019, Sterr2018} and (b) the FB(R) (``front versus back'' for the right ear) derivation \cite{Mikkelsen2019} used in our experiments.}
	\label{fig:cEEGrid}
	\vspace{-0.3cm}
\end{figure}

{\bf Surrey-cEEGrid:} This database \cite{Mikkelsen2019, Sterr2018} was recorded at the University of Surrey using the cEEGrid array \cite{Debener2015,Debener2012}, a novel lightweight flex‐printed electrode strip that fits neatly behind the ear, as illustrated in Figure \ref{fig:cEEGrid} (a). Twenty participants, aged 34.9 ± 13.8 years, had their overnight (about 12 hours) cEEGrid data collected. The PSGs were also recorded in parallel and manual annotation based on the PSG was used as reference for the cEEGrid data \cite{Mikkelsen2019}. Besides two recordings lost due to human error, six recordings were discarded because of excessive artifacts and missing data. A cohort of 12 participants was retained. From the cEEGrid data, the FB(R) (``front versus back'' for the right ear, see Figure \ref{fig:cEEGrid} (b)) EEG derivation, which was the best derivation \cite{Mikkelsen2019}, was obtained and used. We also simulated the two- and three-channel settings by adding the ROC-A2 EOG and CHIN1-CHIN3 channels from the PSG data to the cEEGrid data. Although there exist other EOG and EMG channels, the ROC-A2 EOG and CHIN1-CHIN3 channels were deliberately selected to be different from those of the source domain to maintain the severity of data mismatch.

\setlength\tabcolsep{2.25pt}
\begin{table}[!t]
	\caption{Summary of the employed sleep databases.}
	\scriptsize
	\vspace{-0.25cm}
	\begin{center}
		\begin{tabular}{|>{\arraybackslash}m{0.65in}|>{\centering\arraybackslash}m{0.35in}|>{\centering\arraybackslash}m{0.35in}|>{\centering\arraybackslash}m{0.45in}|>{\centering\arraybackslash}m{0.65in}|>{\centering\arraybackslash}m{0.4in}|>{\centering\arraybackslash}m{0in} @{}m{0pt}@{}}
			\cline{2-6}
			\multicolumn{1}{c|}{} & Num. of subjects & EEG & EOG  & EMG & Data mismatch  & \parbox{0pt}{\rule{0pt}{1ex+\baselineskip}} \\ [0ex]  	
			\cline{1-6}
			MASS & 200 & C4-A1 & ROC-LOC & CHIN1-CHIN2 & - & \parbox{0pt}{\rule{0pt}{0.25ex+\baselineskip}} \\ [0ex]  	
			\cline{1-6}
			Sleep-EDF-SC & 20 & Fpz-Cz & ROC-LOC & - & slight & \parbox{0pt}{\rule{0pt}{0.25ex+\baselineskip}} \\ [0ex]  	
			Sleep-EDF-ST & 22 & Fpz-Cz & ROC-LOC & Submental & slight & \parbox{0pt}{\rule{0pt}{0.25ex+\baselineskip}} \\ [0ex]  	
			Surrey-cEEGrid & 12 & cEEGrid & ROC-A2 & CHIN1-CHIN3 & severe & \parbox{0pt}{\rule{0pt}{0.25ex+\baselineskip}} \\ [0ex]  	
			\cline{1-6}
		\end{tabular}
	\end{center}
	\label{tab:materials}
	\vspace{-0.3cm}
\end{table}

The employed databases and the adopted signals are summarized in Table \ref{tab:materials}. All the signals were downsampled to 100 Hz. The databases were chosen to have the data mismatch between the target domains and the source domain varying from slight level due to the difference in PSG signals used (i.e. Sleep-EDF-SC and Sleep-EDF-ST) to severe level due to completely new electrode placement (i.e. Surrey-cEEGrid).

\vspace{-0.15cm}
\section{The Generic Deep Learning Framework for Sequence-to-Sequence Sleep Staging}
\label{sec:models}

The advent of deep learning has made astonishing progress in automatic sleep staging. First, deep networks are powerful in learning features which outperform and displace traditional handcrafted features. Second, they enable us to achieve automatic sleep stage classification in ways that are impossible for conventional machine-learning algorithms. The sequence-to-sequence sleep staging scheme \cite{Phan2019a} was recently proposed to offer the ability of modelling long-term temporal dependency of sleep data epochs in a deep learning model. Intuitively, a sequence-to-sequence model processes a sequence of multiple consecutive epochs simultaneously and classifies them at once into a sequence of corresponding sleep stages. Here, we frame this scheme into a generic deep learning framework for sequence-to-sequence sleep staging. This framework also sets a potential benchmark to design new models in future work. It is worth noting beforehand that a detail explanation of the network layers and machine learning concepts encountered in the following sections, such as an RNN or a CNN, can be found in \cite{Goodfellow2016}.

\vspace{-0.2cm}
\subsection{The framework}
\label{sec:framework}

Formally, given the input sequence of $L$ consecutive epochs denoted as $(\mathbf{S}_1, \mathbf{S}_2, \ldots, \mathbf{S}_L)$, the sequence-to-sequence sleep staging problem \cite{Phan2019a} is formulated to maximize the conditional probability 
$p(\mathbf{y}_1, \mathbf{y}_2, \ldots, \mathbf{y}_L \,|\, \mathbf{S}_1, \mathbf{S}_2, \ldots, \mathbf{S}_L)$ where $(\mathbf{y}_1, \mathbf{y}_2, \ldots, \mathbf{y}_L)$ represents the sequence of corresponding $L$ one-hot encoding vectors of the ground-truth output labels.

\begin{figure} [!t]
	\centering
	\includegraphics[width=0.65\linewidth]{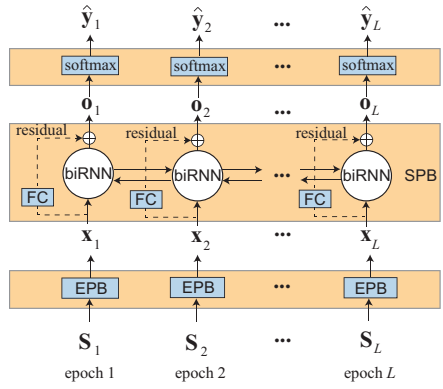}
	\vspace{-0.1cm}
	\caption{The proposed generic deep learning framework for sequence-to-sequence sleep staging which consists of the tied epoch processing block (EPB), the sequence processing block (SPB), and the tied softmax layer.} 
	\label{fig:endtoend_seqtoseq}
	\vspace{-0.25cm}
\end{figure}

The proposed framework are divided into three components, an epoch processing block (EPB), a sequence processing block (SPB), and a softmax layer, as illustrated  in Fig. \ref{fig:endtoend_seqtoseq}. 

{\bf EPB:} Each epoch in the input sequence is presented to the network in some forms of representation (e.g. raw signals \cite{Supratak2017} or time-frequency features \cite{Phan2019a}) and can be single-channel (e.g. EEG or EOG) or multi-channel (e.g. a combination of EEG, EOG, and EMG). The EPB plays the role of an epoch-wise feature learner and extractor. The EPB is common for the PSG epochs in the input sequence and is a sub-network that is trained jointly with other components in an end-to-end manner \cite{Phan2019a}. Via the EPB, an input epoch $\mathbf{S}_l$, $1 \le l \le L$, is transformed into an epoch-wise feature vector $\mathbf{x}_l$. 

{\bf SPB:} The SPB consists of a bidirectional recurrent layer (biRNN) that encodes the sequence of the induced epoch-wise feature vectors $(\mathbf{x}_1, \mathbf{x}_2, \ldots, \mathbf{x}_L)$ into the sequence of output vectors $(\mathbf{o}_1, \mathbf{o}_2, \ldots, \mathbf{o}_L)$. An RNN is a type of deep neural networks that processes an input sequence one element at a time and retain information of all the past elements of the sequence in its hidden state vector \cite{LeCun2015}. A biRNN, on the other hand, consists of two RNN layers of opposite directions to the same input sequence \cite{Schuster1997}. More specifically, the forward and backward recurrent layers of the biRNN iterate over the sequence $(\mathbf{x}_1, \mathbf{x}_2, \ldots, \mathbf{x}_L)$ in opposite directions and compute their forward and backward sequences of hidden state vectors $\mathbf{H}^{\text{f}}~=~(\mathbf{h}^{\text{f}}_1, \mathbf{h}^{\text{f}}_2, \ldots, \mathbf{h}^{\text{f}}_L)$ and $\mathbf{H}^{\text{b}}=(\mathbf{h}^{\text{b}}_1, \mathbf{h}^{\text{b}}_2, \ldots, \mathbf{h}^{\text{b}}_L)$, respectively, where
\begin{align}
\mathbf{h}^{\text{f}}_l &= \mathcal{H}(\mathbf{x}_l\,, \mathbf{h}^{\text{f}}_{l-1}), \label{eq:rnn_hidden_forward} \\
\mathbf{h}^{\text{b}}_l &= \mathcal{H}(\mathbf{x}_l\,, \mathbf{h}^{\text{b}}_{l+1}), \mbox{~} 1 \le l \le L.
\label{eq:rnn_hidden_backward}
\end{align}
In (\ref{eq:rnn_hidden_forward}) and (\ref{eq:rnn_hidden_backward}),  $\mathcal{H}$ denotes the hidden layer function of the biRNN and can be realized by either Long Short-Term Memory (LSTM) \cite{Hochreiter1997} or Gated Recurrent Unit (GRU) \cite{Cho2014}, two most popular RNN variants. The sequence of output vectors $(\mathbf{o}_1, \mathbf{o}_2, \ldots, \mathbf{o}_L)$ is then computed:
\begin{align}
\mathbf{o}_l &= \mathbf{W}_{ho}[\mathbf{h}^{\text{b}}_l \oplus \mathbf{h}^{\text{f}}_l] + \mathbf{b}_{o}, \mbox{~} 1 \le l \le L,
\label{eq:seq_rnn_output}
\end{align}
where $\oplus$ represents vector concatenation. In (\ref{eq:seq_rnn_output}), $\mathbf{W}_{ho}$ denotes a learnable weight matrix and $\mathbf{b}_{o}$ denotes a learnable bias. The (long-term) dependency of the input epochs are expected to be modelled by the biRNN layer and the output vectors $\mathbf{o}_l$, $1 \le l \le L$ are expected to encode sequence-level information. A residual connection can be optionally used to integrate epoch-wise features $\mathbf{x}_l$ and sequence-wise features $\mathbf{o}_l$ and, hence, enables the network to explore their combination in the classification stage. The fully-connected layer (FC) of the residual connection is to convert $\mathbf{x}_l$ into another vector having its size compatible to $\mathbf{o}_l$ for a proper residual combination. All the residual connections also share their parameters. 

{\bf Softmax:} The classification is carried out by the shared softmax layer to yield the output sequence of sleep stage probabilities $(\mathbf{\hat{y}}_1, \mathbf{\hat{y}}_2, \ldots,\mathbf{\hat{y}}_L)$ from the sequence of output vectors $(\mathbf{o}_1, \mathbf{o}_2, \ldots, \mathbf{o}_L)$. Different from SeqSleepNet in \cite{Phan2019a} and DeepSleepNet in \cite{Supratak2017}, we use a common softmax layer for classification at all indices $1, 2, \ldots, L$ to reduce the number of network parameters rather than one separate softmax layer at each of the indices.
A network that adheres to this framework can be trained to minimize the sequence classification loss over $N$ training sequences in the training data:
\begin{align}
E(\bm{\theta}) = -\frac{1}{L}\sum_{n=1}^{N}\sum_{l=1}^{L} \mathbf{y}_l\log\left(\mathbf{\hat{y}}_l\left(\bm{\theta}\right)\right) + \frac{\lambda}{2}\|\bm{\theta}\|^2_2.
\label{eq:sequence_loss}
\end{align}
Here, $\bm{\theta}$ represents the network parameters and $\lambda$ denotes the hyper-parameter that trades off the error terms and the $\ell_2$-norm regularization term.

\vspace{-0.2cm}
\subsection{The derived networks}
From the framework presented in Section \ref{sec:framework}, we develop two networks as the base models for transfer learning:

\begin{figure} [!t]
	\centering
	\includegraphics[width=0.875\linewidth]{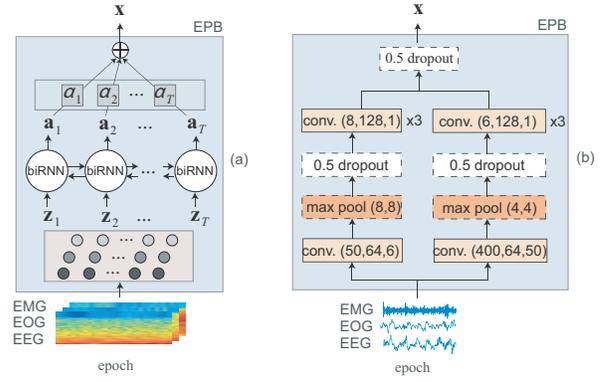}
	\vspace{-0.1cm}
	\caption{Illustration of the EPBs of (a) SeqSleepNet+ and (b) DeepSleepNet+. The former relies on an attentional biRNN coupled with filterbank layers. The latter is a two-branch deep CNN.}
	\label{fig:EFE}
	\vspace{-0.25cm}
\end{figure}

{\bf SeqSleepNet+:} This network is similar to SeqSleepNet presented in \cite{Phan2019a}, except that a common softmax layer is used at all indices of the input sequence. Hence, SeqSleepNet+ is more compact than SeqSleepNet \cite{Phan2019a}. The network receives the log-scale time-frequency representation \cite{Phan2019a} as input. The time-frequency image is normalized to zero-mean and unit standard deviation. In case of multi-channel, the channel-wise image features are stacked as a multi-channel image. The network's EPB is realized by \emph{filterbank} layers \cite{phan2018c, Phan2019a}, one for each input image channel for preprocessing purpose, followed by an attentional biRNN as illustrated in Figure \ref{fig:EFE}(a). Note that this EBP's biRNN should not be confused with the SBP's biRNN in Fig. \ref{fig:endtoend_seqtoseq}. Both the EPB's biRNN and the SPB's biRNN of the network are implemented by a GRU cell \cite{Cho2014} with recurrent batch normalization \cite{Cooijmans2016}. There is no residual connection (cf. Figure \ref{fig:endtoend_seqtoseq}) in the SPB of this network.

{\bf DeepSleepNet+:} This network is inherited from DeepSleepNet \cite{Supratak2017} and its end-to-end variant \cite{Phan2019a}, except for the common softmax used at all indices of the input sequence. The network receives raw signals as input. When the input are composed of multiple signals, the raw signal are stacked to form a multi-channel input. The network's EPB is composed of two deep CNNs organized in two branches with 4 convolutional layers each as illustrated in Figure \ref{fig:EFE}(b). A CNN is a type of deep neural networks designed to efficiently process data that come in the form of multiple arrays \cite{LeCun2015}, such as one-dimensional signals in this case. A CNN features local connections, shared weights, and pooling to learn translation-invariant features from the input.  The convolutional kernels in the two branches are purposely designed to have different sizes so that they can learn features at both fine and coarse temporal resolutions. Each convolutional layer is associated with batch normalization \cite{Ioffe2015} and Rectified Linear Units (ReLU) activation \cite{Nair2010}. The SPB's biRNN relies on the LSTM cell  \cite{Hochreiter1997} and is designed to have two bidirectional LSTM layers, one stacked on top of the other. In addition, the SPB makes use of the residual connection.

As the two networks inherits SeqSleepNet's and DeepSleepNet's architecture's, respectively, they are divergent in their inputs, EPB, and SPB components \cite{Phan2019d}. Therefore, these differences suggest discrepant behaviors during transfer learning.

\vspace{-0.15cm}
\section{Transfer Learning Scenarios for Automatic Sleep Staging on Small Cohorts}

Formally, let $\mathcal{D}_S=\{\mathcal{X}_S, \mathcal{Y}_S\}$ denote the source domain with the feature space $\mathcal{X}_S$ and the label space $\mathcal{Y}_S$. In addition, let $\mathcal{T}_S$ denote the task in the source domain with the source conditional probability distributions $P(\mathbf{y}_S\,|\,\mathbf{x}_S)$, where $\mathbf{x}_S \in \mathcal{X}_S$ and $\mathbf{y}_S \in \mathcal{Y}_S$. Similarly, $\mathcal{D}_T=\{\mathcal{X}_T, \mathcal{Y}_T\}$ denotes the target domain with the feature space $\mathcal{X}_T$ and the label space $\mathcal{Y}_T$. $\mathcal{T}_S$ denotes the task in the target domain with the conditional probability distributions $P(\mathbf{y}_T\,|\,\mathbf{x}_T)$, where $\mathbf{x}_T \in \mathcal{X}_T$ and $\mathbf{y}_T \in \mathcal{Y}_T$, respectively. The objective of transfer learning is to improve learning $P(\mathbf{y}_T\,|\,\mathbf{x}_T)$ with information gained from $\mathcal{D}_S$  and $\mathcal{T}_S$ where $\mathcal{D}_S \neq \mathcal{D}_T$ or  $\mathcal{T}_S \neq \mathcal{T}_T$ \cite{Pan2010}. In our case, $\mathcal{T}_S \equiv \mathcal{T}_T$, as we aim at performing sleep staging with the same set of sleep stages in both the source and target domains. Transfer learning \cite{Pan2010} relaxes the hypothesis that the training data must be  identically distributed as the test data. Therefore, it is useful to deal with data mismatch and holds promise to leverage the large amount of available data to overcome the problem of having insufficient training data in small cohort studies.

In the present context, a model (e.g. SeqSleepNet+ or DeepSleepNet+) is firstly trained in the source domain  and then finetuned in the target domain to complete knowledge transfer as illustrated in Figure \ref{fig:sleep_transfer}. Without loss of generality, the pretraining process is to minimize the loss $L_S$ over the source-domain data, resulting in the model parameter $\bm{\theta}$:
\begin{align}
\argmin_{\bm{\theta}} &= \sum_{(\mathbf{x},\mathbf{y}) \in \mathcal{D}_S}L_{S}\left(\mathbf{x}, P(\mathbf{y}\,|\,\mathbf{x}), P_{\bm{\theta}}\left(\mathbf{y}\,|\,\mathbf{x}\right)\right).
\label{eq:seq_rnn_output_pretrain}
\end{align}
The pretrained model is considered as a starting point in the target domain. To accomplish transfer learning, a subset of the pretrained network's parameter $\bm{\theta'} \subseteq \bm{\theta}$ is finetuned (i.e. further trained) with the target-domain data while the rest $\bm{\theta} \backslash \bm{\theta'}$ remains unchanged (i.e. being reused):
\begin{align}
\argmin_{\bm{\theta'} \subseteq \bm{\theta}} &= \sum_{(\mathbf{x},\mathbf{y}) \in \mathcal{D}_T}L_{T}\left(\mathbf{x}, P(\mathbf{y}\,|\,\mathbf{x}), P_{\bm{\theta}}\left(\mathbf{y}\,|\,\mathbf{x}\right)\right).
\label{eq:seq_rnn_output_finetune}
\end{align}
When $\bm{\theta'} = \bm{\theta}$, the entire pretrained network is finetuned in the target domain. In contrast, when $\bm{\theta'} = \emptyset$, no finetuning happens and the pretrained network is directly used in the target domain.

In order to study the influence of finetuning different components of a pretrained SeqSleepNet+ and DeepSleepNet+ to the sleep staging performance on the target domains, we examine four \emph{finetuning strategies} corresponding to different component combinations: all, EPB+softmax, SPB+softmax, and softmax. The parameter subsets corresponding to these combinations will be adapted with the target-domain data while the rest remains fixed. The case in which the pretrained network is directly used in the target domain without finetuning is considered as a baseline. The finetuning strategies are carried out to study the following \emph{transfer learning scenarios}:

\begin{figure} [!t]
	\centering
	\includegraphics[width=.5\linewidth]{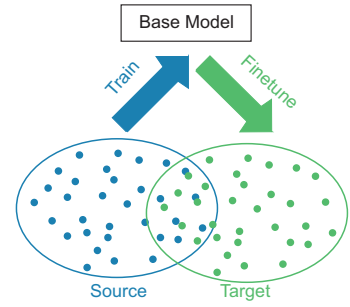}
	\vspace{-0.15cm}
	\caption{Transfer learning from a source domain to a target domain. The base model is trained using the source-domain data, and then finetuned on the target-domain data to complete knowledge transfer.}
	\label{fig:sleep_transfer}
	\vspace{-0.25cm}
\end{figure}

{\bf EEG$\cdot$EOG$\cdot$EMG$\mapsto$EEG$\cdot$EOG$\cdot$EMG:} Apart from brain activities, sleep also involves eye movements and muscular activities at different levels. For instance, Rapid Eye Movement (REM) stage usually associates with rapid eye movements and high muscular activities are usually seen during the Awake stage. Therefore, EOG and EMG are valuable additional sources, complementing EEG in the automatic sleep staging task \cite{Stephansen2018,Mikkelsen2018, Chambon2018, Andreotti2018, Phan2019b}. We study this three-channel EEG$\cdot$EOG$\cdot$EMG transfer learning scenario when all EEG, EOG, and EMG are available in a target domain (i.e. in case of Sleep-EDF-ST and Surrey-cEEGrid).

{\bf EEG$\cdot$EOG$\mapsto$EEG$\cdot$EOG:} This scenario assumes the unavailability of EMG and examines two-channel EEG·EOG transfer learning. Different from the three-channel case, we are able to study this scenario across all the adopted target domains as they all have full EEG and EOG recordings available.

{\bf EEG$\mapsto$EEG:} This scenario explores single-channel EEG transfer learning. Automatic sleep staging with single-channel EEG is prevalent in literature \cite{Koley2012, Tsinalis2016b, phan2018c, phan2018d, Kuo2011, Phan2013}. Without the augmentation from EOG and EMG, this single-channel setting usually results in a lower performance compared to those of the multi-channel ones; however, it is desirable due to the simple configuration. It is particularly useful for sleep monitoring applications with mobile EEG devices \cite{Mikkelsen2019, Sterr2018}.

{\bf EOG$\mapsto$EOG:} In general, EOG signals contain rich information from multiple sources, including ocular activity, frontal EEG activity, and EMG from cranial and eye muscles \cite{Olesen2016}. They are, therefore, promising alternatives for EEG in single-channel sleep staging. In addition, due to the ease of electrode placements, it would be ideal for home-based sleep monitoring applications with wearable devices \cite{Mikkelsen2019, Sterr2018}. Despite their potential, EOG signals have been mainly used as secondary modality in multi-channel sleep staging studies \cite{Olesen2016, Liang2015}. With this scenario, we aim to exploit standalone EOG and deep transfer learning on this secondary modality to examine whether its performance is comparable to that using the primary EEG in single-channel sleep staging. 

{\bf EEG$\mapsto$EOG:} As an extension of the EOG$\mapsto$EOG scenario, this cross-modality transfer learning scenario investigates whether a base model trained on EEG in the source domain can be transferred to EOG in the target domain and if its performance is comparable to the same-domain EOG$\mapsto$EOG transfer learning scenario. If the answers to these questions are true, instead of modality-specific pretrained models, a single model pretrained solely on EEG can serve as a generic model for single-channel transfer learning regardless the modality of the target domain.

Apart from the data mismatch caused by the differences in recording devices and/or electrode placements in case of the same-modality scenarios (i.e. EEG$\cdot$EOG$\cdot$EMG$\mapsto$EEG$\cdot$EOG$\cdot$EMG, EEG$\cdot$EOG$\mapsto$EEG$\cdot$EOG, EEG$\mapsto$EEG, and EOG$\mapsto$EOG), heavy data mismatch is expected in case of the cross-modality EEG$\mapsto$EOG scenario when the base models are trained with EEG data in the source domain is transferred to EOG data in the target domains. On the one hand, with the same-modality scenarios, we aim to show that even when the source domain and the target domains are of the same modalities, transfer learning is still necessary. On the other hand, the cross-modality scenario is to emphasize that transfer learning is efficient in tackling heavy data mismatch to transfer knowledge from the source domain to the target domains.

\vspace{-0.2cm}
\section{Experiments}
\subsection{Experimental Setup}
SeqSleepNet+ and DeepSleepNet+ were pretrained using the data from the entire 200 subjects of the MASS database (i.e. the source domain) and then finetuned in the target domains. To evaluate the efficiency of transfer learning on sleep staging in the target domains, cross-validation was conducted. Leave-one-out cross-validation was conducted for Sleep-EDF-SC (20 subjects), and Surrey-cEEGrid (12 subjects) while 11-fold cross-validation was performed for Sleep-EDF-ST (22 subjects) to have an equal number of test subjects (i.e. 2 subjects) in each cross-validation fold. At each iteration of cross-validation, a number of subjects were randomly selected and left out for validation purpose, i.e. for early stopping the finetuning process, (4 for Sleep-EDF-SC and Sleep-EDF-ST and 2 for Surrey-cEEGrid). The performance over all cross-validation folds was then calculated. 

\vspace{-0.25cm}
\subsection{Network Parameters}
Both SeqSleepNet+ and DeepSleepNet+ were implemented using \emph{Tensorflow} \cite{Abadi2016}. The networks were parametrized similar to SeqSleepNet and DeepSleepNet in our previous work \cite{Phan2019a}. We experimented with the input sequence length $L = 20$ epochs as this value is a reasonable choice for these sequence-to-sequence models \cite{Phan2019a}. The sequences were sampled from the training recordings with a hop size of one epoch for network training and finetuning. During testing, the test sequences were also shifted by one epoch, resulting in an ensemble of $L$ classification decisions at each epoch of a test recordings. A probabilistic aggregation step similar to \cite{Phan2019a} was carried out to fuse the decision ensemble into the final decision.

In the source domain, the networks were pretrained with the MASS database for 10 training epochs with a minibatch size of 32 sequences. For transfer learning, the pretrained networks were further finetuned on each target-domain databases for 10 finetuning epochs. The finetuning process was stopped early when no accuracy improvement was seen on the validation subjects for 50 finetuning steps. Both network training and finetuning were performed using \emph{Adam} optimizer, an optimization algorithm proposed in \cite{Kingma2015} for training deep neural networks. This optimizer leverages the power of adaptive learning rates methods to find individual learning rates for each parameter of the network. The initial learning rate of Adam optimizer was set to $10^{-4}$.

\setlength\tabcolsep{2.25pt}
\begin{table}[!b]
	\caption{Sleep staging performance on the source domain (i.e. the MASS database).}
	\vspace{-0.3cm}
	\footnotesize
	\begin{center}
		\begin{tabular}{|>{\arraybackslash}m{0.8in}|>{\centering\arraybackslash}m{0.25in}|>{\centering\arraybackslash}m{0.3in}|>{\centering\arraybackslash}m{0.25in}|>{\centering\arraybackslash}m{0.25in}|>{\centering\arraybackslash}m{0.3in}|>{\centering\arraybackslash}m{0.25in}|>{\centering\arraybackslash}m{0in} @{}m{0pt}@{}}
			\cline{1-7}
			\multirow{2}{*}{} & \multicolumn{3}{c|}{SeqSleepNet} & \multicolumn{3}{c|}{DeepSleepNet} & \parbox{0pt}{\rule{0pt}{1ex+\baselineskip}} \\ [0ex]  	
			
			\cline{2-7}
			Input & Acc. & MF1 & $\kappa$ & Acc. & MF1 & $\kappa$ & \parbox{0pt}{\rule{0pt}{1ex+\baselineskip}} \\ [0ex]  	
			\cline{1-7}
			EEG$\cdot$EOG$\cdot$EMG & $87.0$ & $83.3$ & $0.815$ & $86.5$ & $82.4$ & $0.807$ &  \parbox{0pt}{\rule{0pt}{0.25ex+\baselineskip}} \\ [0ex]  	
			EEG$\cdot$EOG & $86.5$ & $82.4$ & $0.808$ & $85.9$ & $81.6$ & $0.799$ &  \parbox{0pt}{\rule{0pt}{0ex+\baselineskip}} \\ [0ex]  	
			EEG & $84.5$  & $79.8$ & $0.778$& $84.3$ & $79.7$ & $0.777$ & \parbox{0pt}{\rule{0pt}{0ex+\baselineskip}} \\ [0ex]  	
			EOG & $83.9$ & $79.1$ & $0.769$ & $83.7$  & $78.9$ & $0.767$& \parbox{0pt}{\rule{0pt}{0ex+\baselineskip}} \\ [0ex]  	
			\cline{1-7}
		\end{tabular}
	\end{center}
	\label{tab:performance_source}
	\vspace{-0.3cm}
\end{table}

\vspace{-0.2cm}
\subsection{Experimental Results}
\label{ss:experimental_results}
\subsubsection{Performance on the source domain}

It is first worth assessing SeqSleepNet+ and DeepSleepNet+ on the source domain to see how well they perform on a large number of subjects across the input spectrum. 
To this end, we conducted 10-fold cross-validation on the source domain. At each iteration, 180 subjects were used for training, 10 for validation, and 10 for testing. The results of the cross-validation folds were finally pooled to calculate the overall metrics, including accuracy, macro F1-score, and Cohen's kappa ($\kappa$). The obtained performance with different inputs are shown in Table \ref{tab:performance_source}. Firstly, the results in the table confirm the benefit of using EOG and EMG to complement EEG in the automatic sleep staging task as their presence lead to performance improvement. Secondly, with the sequence-to-sequence framework, the performance obtained by the secondary EOG is just marginally lower than that of the primary EEG, evidenced by both SeqSleepNet+ and DeepSleepNet+. This suggests that EOG can be used as a standalone modality similar to EEG when a single channel is used. 

\begin{figure} [!t]
	\centering
	\includegraphics[width=0.9\linewidth]{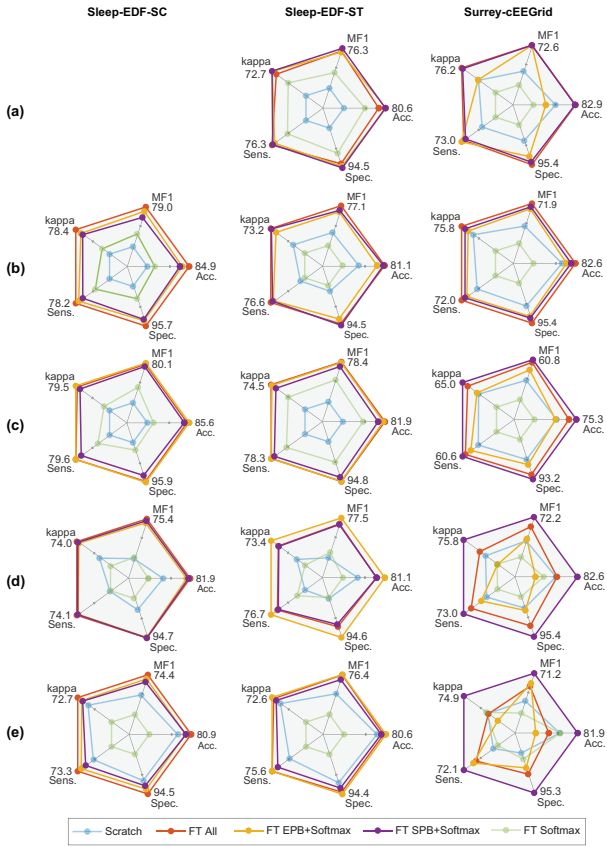}
	\caption{Performance patterns obtained by finetuning the pretrained SeqSleepNet+ with different finetuning strategies in comparison with those of the SeqSleepNet+ scratch model. (a) EEG$\cdot$EOG$\cdot$EMG$\mapsto$EEG$\cdot$EOG$\cdot$EMG, (b) EEG$\cdot$EOG$\mapsto$EEG$\cdot$EOG, (c) EEG$\mapsto$EEG, (d) EOG$\mapsto$EOG, and (e) EEG$\mapsto$EOG.}
	\label{fig:seqsleepnet_finetune}
	\vspace{-0.35cm}
\end{figure}

\begin{figure} [!t]
	\centering
	\includegraphics[width=0.89\linewidth]{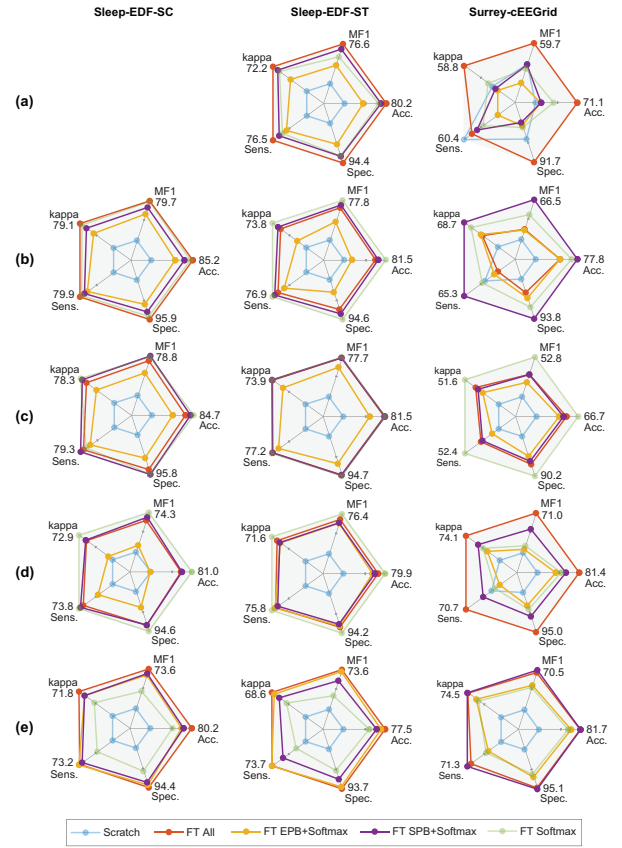}
	\caption{Performance patterns obtained by finetuning the pretrained DeepSleepNet+ with different finetuning strategies in comparison with those of the DeepSleepNet+ scratch model. (a) EEG$\cdot$EOG$\cdot$EMG$\mapsto$EEG$\cdot$EOG$\cdot$EMG, (b) EEG$\cdot$EOG$\mapsto$EEG$\cdot$EOG, (c) EEG$\mapsto$EEG, (d) EOG$\mapsto$EOG, and (e) EEG$\mapsto$EOG.}
	\label{fig:deepsleepnet_finetune}
	\vspace{-0.35cm}
\end{figure}

\subsubsection{The effect of transfer learning on the target domains}
\label{sssec:transfer_learning_effects}
Figures \ref{fig:seqsleepnet_finetune} and \ref{fig:deepsleepnet_finetune} give an overall picture on the performance obtained by SeqSleepNet+ and DeepSleepNet+ on the target domains with respect to different finetuning strategies and compared to the model trained from scratch using the target-domain data only.
The two networks show noticeably varying patterns on the transfer learning results. 

On the one hand, SeqSleepNet+'s results in Figure \ref{fig:seqsleepnet_finetune} reveal that, while finetuning the softmax layer alone leads to better performance than that of the scratch model in some cases, it is essential to additionally finetune the feature-learning parts of the network, either the EPB for epoch-level feature learning or the SPB for sequence-level feature learning, or both. This pattern exists across all finetuning cases in the figure. This suggests that the features learned by SeqSleepNet+ in the source domain are slightly different from those in the target domain. This is reasonable due to the data mismatch between the source and target domains.

On the other hand, DeepSleepNet+'s finetuning results expose diverging patterns as shown in Figure \ref{fig:deepsleepnet_finetune}. Finetuning the softmax layer alone results in comparable, or even better, performance than other finetuning strategies in some cases (such as the EEG$\mapsto$EEG scenarios) whereas its results are largely belittled in other cases (such as EEG$\mapsto$EEG scenarios). This suggests that, when the signals are of the same modality, the features learned from the source domain's raw signals persist in the target domain and only their combinations need to be adapted in the target domains. In contrast, when the signals are from different modalities, additional finetuning the feature learning parts (i.e. the EPB or the SPB or both) is necessary. It, however, should be emphasized that persistence of the learned features across the source and target domains does not necessarily mean good generalization as DeepSleepNet+'s finetuning results are inferior to those of its counterpart, SeqSleepNet+ (see Table \ref{tab:performance_comaprison}).

\begin{figure} [!t]
	\centering
	\includegraphics[width=0.95\linewidth]{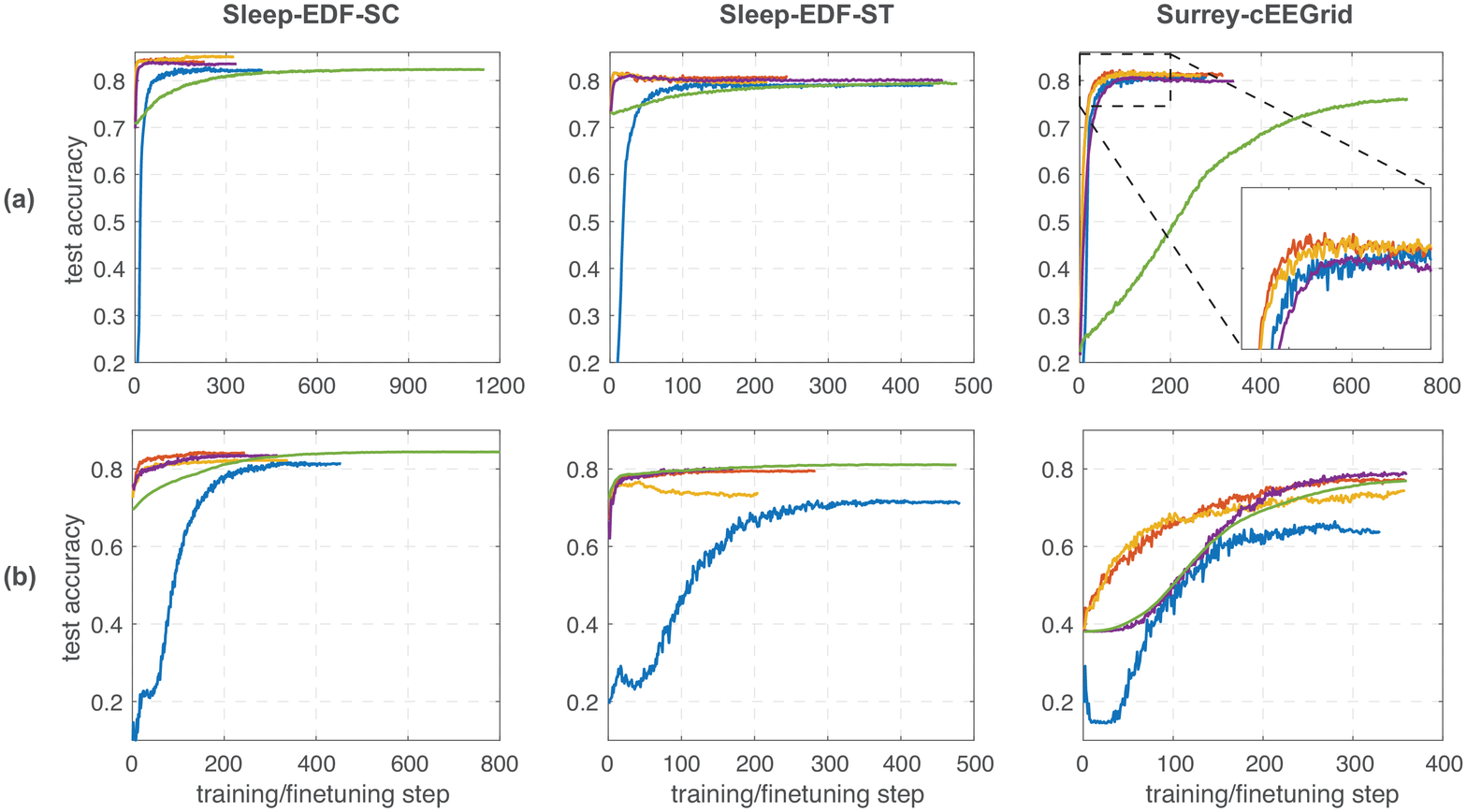}
	\vspace{-0.1cm}
	\caption{The test accuracy during the finetuning/training. Each curve was averaged over all the cross-validation folds. (a) SeqSleepNet+ and (b) DeepSleepNet+.}
	\label{fig:learning_curve}
	\vspace{-0.25cm}
\end{figure}

Despite their different behaviors in finetuning, both SeqSleepNet+ and DeepSleepNet+ meet the transfer learning's expectation. Compared to the network trained from scratch using the target-domain data only, transfer learning consistently results in improvements across different network types, the target domains, and the transfer learning scenarios. The benefits of transfer learning is further evidenced by contrasting the learning curves of the finetuned models and the scratch models. Taking the two-channel EEG$\cdot$EOG$\mapsto$EEG$\cdot$EOG scenario as an example (see Figure \ref{fig:learning_curve}), the learning curves were recorded on the test data during finetuning and training, respectively. Each learning curve was averaged over all cross-validation folds. As the learning curves' lengths vary across different folds due to early stopping, those with shorter length than the maximum one were padded to the maximum length before averaging. SeqSleepNet+'s learning curves show better generalization and faster convergence of the finetuned models (except the softmax-only finetuning strategy) compared to their scratch opponents. Similar motifs are observed in DeepSleepNet+'s learning curves; however, the softmax-only finetuning strategy shows a comparable generalization to other strategies (although slower convergence). These findings consolidate the explanation for the finetuning results in Figures \ref{fig:seqsleepnet_finetune} and \ref{fig:deepsleepnet_finetune}.

\setlength\tabcolsep{2.25pt}
\begin{table*}[!t]
	\caption{Performance comparison between the proposed transfer-learning systems, and the baseline systems (i.e. the scratch models and the direct-transfer models, in italic font), and previous works. FT and DT are abbreviated for ``finetuning'' and ``direct transfer'', respectively. It should be noted that the comparison may not be completely compatible due to differences in experimental setup: $^\ast$the transfer learning approach was personalized finetuning; $^\dagger$30 minutes of data (mainly Wake epochs) before and after in-bed duration were included, therefore, the results are likely biased towards Wake stage; $^\ddagger$the evaluation was not subject-independent \cite{Phan2019b}; $^\diamond$the number of subjects is different from that in our experiments.} 
	\scriptsize
	\vspace{-0.25cm}
	\begin{center}
		\begin{tabular}{|>{\arraybackslash}m{0.1in}|>{\arraybackslash}m{1.2in}|>{\centering \arraybackslash}m{0.3in}|>{\centering\arraybackslash}m{0.25in}|>{\centering\arraybackslash}m{0.25in}|>{\centering\arraybackslash}m{0.3in}|>{\centering\arraybackslash}m{0.25in}|>{\centering\arraybackslash}m{0.25in}|>{\centering\arraybackslash}m{0.3in}|>{\centering\arraybackslash}m{0.25in}|>{\centering\arraybackslash}m{0.25in}|>{\centering\arraybackslash}m{0.3in}|>{\centering\arraybackslash}m{0.25in}|>{\centering\arraybackslash}m{0.25in}|>{\centering\arraybackslash}m{0.3in}|>{\centering\arraybackslash}m{0.25in}|>{\centering\arraybackslash}m{0.25in}|>{\centering\arraybackslash}m{0.3in}|>{\centering\arraybackslash}m{0in} @{}m{0pt}@{}}
			\cline{2-18}
			\multicolumn{1}{c|}{} & \multirow{2}{*}{~~~~~~~~System}& \multirow{2}{*}{\makecell{Transfer\\learning}} & \multicolumn{3}{c|}{\makecell{~~~EEG$\cdot$EOG$\cdot$EMG\\$\mapsto$EEG$\cdot$EOG$\cdot$EMG}} & \multicolumn{3}{c|}{EEG$\cdot$EOG$\mapsto$EEG$\cdot$EOG} & \multicolumn{3}{c|}{EEG$\mapsto$EEG} & \multicolumn{3}{c|}{EOG$\mapsto$EOG} & \multicolumn{3}{c|}{EEG$\mapsto$EOG} & \parbox{0pt}{\rule{0pt}{1ex+\baselineskip}} \\ [0ex]  	
			\cline{4-18}
			\multicolumn{1}{c|}{} &  & & Acc. & MF1 & $\kappa$ & Acc. & MF1 & $\kappa$ & Acc. & MF1 & $\kappa$ & Acc. & MF1 & $\kappa$ & Acc. & MF1 & $\kappa$ &  \parbox{0pt}{\rule{0pt}{1ex+\baselineskip}} \\ [0ex]  	
			\cline{1-18}
			
			\multirow{21}{*}{\begin{sideways}{\bf Sleep-EDF-SC~~~}\end{sideways}} & \bf FT SeqSleepNet+ & Yes & & & & $\bm{84.3}$ & $\bm{77.7}$ & $\bm{0.776}$ & $\bm{85.2}$ & $\bm{79.6}$ & $\bm{0.789}$ & $\bm{81.7}$ & $\bm{75.1}$ & $\bm{0.737}$ & $\bm{80.0}$ & $\bm{72.3}$ & $\bm{0.709}$ & \parbox{0pt}{\rule{0pt}{0ex+\baselineskip}} \\ [0ex]  	
			
			& \bf FT DeepSleepNet+  & Yes & $-$ & $-$ & $-$ & $\bm{84.6}$ & $\bm{79.0}$ & $\bm{0.782}$ & $\bm{84.4}$ & $\bm{78.8}$ & $\bm{0.781}$ & $\bm{79.8}$ & $\bm{73.4}$ & $\bm{0.713}$ & $\bm{79.4}$ & $\bm{72.8}$ & $\bm{0.707}$ & \parbox{0pt}{\rule{0pt}{0ex+\baselineskip}} \\ [0ex]  	
			
			& \it DT SeqSleepNet+ & Yes & $-$ & $-$ & $-$ & $72.0$ & $62.1$ & $0.601$ & $81.2$ & $74.6$ & $0.733$ & $67.2$ & $59.1$ & $0.530$ & $51.1$ & $42.5$ & $0.300$
			& \parbox{0pt}{\rule{0pt}{0ex+\baselineskip}} \\ [0ex]  	
			
			& \it DT DeepSleepNet+ & Yes & $-$ & $-$ & $-$ & $70.2$ & $59.8$ & $0.586$ & $74.2$ & $66.9$ & $0.651$ & $54.1$ & $41.9$ & $0.396$ & $39.7$ & $35.8$ & $0.235$ & \parbox{0pt}{\rule{0pt}{0ex+\baselineskip}} \\ [0ex]  	
			
			& \it Scratch SeqSleepNet+  & No & $-$ & $-$ & $-$ & $82.2$ & $74.2$ & $0.744$ & $82.2$ & $74.1$ & $0.746$ & $78.5$ & $68.3$ & $0.688$ & $78.5$ & $68.3$ & $0.688$ &  \parbox{0pt}{\rule{0pt}{0ex+\baselineskip}} \\ [0ex]  	
			
			&  \it Scratch DeepSleepNet+ & No & $-$ & $-$ & $-$ & $81.9$ & $75.2$ & $0.744$ & $80.8$ & $74.2$ & $0.731$ & $75.9$ & $66.9$ & $0.652$ & $75.9$ & $66.9$ & $0.652$  & \parbox{0pt}{\rule{0pt}{0ex+\baselineskip}} \\ [0ex]  	
			
			& Personalized Deep CNN$^\ast$ \cite{Mikkelsen2018} & Yes & $-$ & $-$ & $-$ &  $84.0$  & $-$ & $-$ & $-$  & $-$ & $-$ & $-$  & $-$ & $-$ & $-$  & $-$ & $-$ & \parbox{0pt}{\rule{0pt}{0ex+\baselineskip}} \\ [0ex]  	
			
			& VGG-FT \cite{Vilamala2017} & Yes & $-$ & $-$ & $-$ &  $-$  & $-$ & $-$& $80.3$  & $-$ & $-$ & $-$  & $-$ & $-$ & $-$  & $-$ & $-$ & \parbox{0pt}{\rule{0pt}{0ex+\baselineskip}} \\ [0ex]  	
			
			& VGG-FE \cite{Vilamala2017} & Yes & $-$ & $-$ & $-$ &  $-$  & $-$ & $-$& $76.3$  & $-$ & $-$ & $-$  & $-$ & $-$ & $-$  & $-$ & $-$ & \parbox{0pt}{\rule{0pt}{0ex+\baselineskip}} \\ [0ex]  	
			& ResNet \cite{Andreotti2018} & Yes & $-$ & $-$ & $-$ &  $76.8$  & $-$ & $-$ & $-$  & $-$ & $-$ & $-$  & $-$ & $-$ & $-$  & $-$ & $-$ & \parbox{0pt}{\rule{0pt}{0ex+\baselineskip}} \\ [0ex]  	
			
			& U-time \cite{Perslev2019}$^\dagger$ & No & $-$ & $-$ & $-$ &  $-$  & $-$ & $-$& $-$  & $79.0$ & $-$ & $-$  & $-$ & $-$ & $-$  & $-$ & $-$ & \parbox{0pt}{\rule{0pt}{0ex+\baselineskip}} \\ [0ex]  	
			
			& IITNet \cite{Back2019}$^\dagger$ & No & $-$ & $-$ & $-$ &  $-$  & $-$ & $-$& $84.0$  & $77.7$ & $0.78$ & $-$  & $-$ & $-$ & $-$  & $-$ & $-$ & \parbox{0pt}{\rule{0pt}{0ex+\baselineskip}} \\ [0ex]  	
			
			& DeepSleepNet$^\dagger$ \cite{Supratak2017} & No & $-$ & $-$ & $-$ & $-$  & $-$ & $-$ &  $82.0$  & $76.9$ & $0.760$ & $-$  & $-$ & $-$ & $-$  & $-$ & $-$ & \parbox{0pt}{\rule{0pt}{0ex+\baselineskip}} \\ [0ex]  	
			
			& Multitask 1-max CNN \cite{Phan2019b} & No & $-$ & $-$ & $-$ & $82.3$  & $74.7$ & $0.750$ & $81.9$ & $73.8$ & $0.740$ & $-$  & $-$ & $-$ & $-$  & $-$ & $-$ & \parbox{0pt}{\rule{0pt}{0ex+\baselineskip}} \\ [0ex]  	
			
			& 1-max CNN \cite{phan2018c} & No & $-$ & $-$ & $-$ & $-$  & $-$ & $-$ & $79.8$ & $72.0$ & $0.720$ & $-$  & $-$ & $-$ & $-$  & $-$ & $-$ & \parbox{0pt}{\rule{0pt}{0ex+\baselineskip}} \\ [0ex]  	
			
			& Attentional RNN \cite{phan2018d} & No & $-$ & $-$ & $-$ &  $-$  & $-$ & $-$& $79.1$  & $69.8$ & $0.700$ & $-$  & $-$ & $-$ & $-$  & $-$ & $-$ & \parbox{0pt}{\rule{0pt}{0ex+\baselineskip}} \\ [0ex]  	
			
			& Deep auto-encoder \cite{Tsinalis2016b} & No & $-$ & $-$ & $-$ &  $-$  & $-$ & $-$& $78.9$  & $73.3$ & $-$ & $-$  & $-$ & $-$ & $-$  & $-$ & $-$ & \parbox{0pt}{\rule{0pt}{0ex+\baselineskip}} \\ [0ex]  	
			
			& Deep CNN \cite{Tsinalis2016} & No & $-$ & $-$ & $-$ &  $-$  & $-$ & $-$& $74.8$  & $69.8$ & $-$ & $-$  & $-$ & $-$ & $-$  & $-$ & $-$ & \parbox{0pt}{\rule{0pt}{0ex+\baselineskip}} \\ [0ex]  	
			
			
			& Decision trees \cite{Aboalayon2016}$^\ddagger$ & No & $-$ & $-$ & $-$ &  $-$  & $-$ & $-$& $93.1$  & $-$ & $-$ & $-$  & $-$ & $-$ & $-$  & $-$ & $-$ & \parbox{0pt}{\rule{0pt}{0ex+\baselineskip}} \\ [0ex]  	
			
			& $k$-NN \cite{Rodriguez-Sotelo2014}$^\ddagger$ & No & $-$ & $-$ & $-$ &  $-$  & $-$ & $-$& $80.0$  & $-$ & $-$ & $-$  & $-$ & $-$ & $-$  & $-$ & $-$ & \parbox{0pt}{\rule{0pt}{0ex+\baselineskip}} \\ [0ex]  	
			
			& GMM \cite{Munk2018}$^\ddagger$ & No & $-$ & $-$ & $-$ & $-$  & $-$ & $-$& $73.3$  & $-$ & $-$ & $-$  & $-$ & $-$ & $-$  & $-$ & $-$ & \parbox{0pt}{\rule{0pt}{0ex+\baselineskip}} \\ [0ex]  	
			
			\cline{1-18}

			\multirow{7}{*}{\begin{sideways}{\bf Sleep-EDF-ST~~~}\end{sideways}} & \bf FT SeqSleepNet+ & Yes & $\bm{80.6}$ & $\bm{76.2}$ & $\bm{0.727}$ & $\bm{81.0}$ & $\bm{76.7}$ & $\bm{0.732}$ & $\bm{81.0}$ & $\bm{77.5}$ & $\bm{0.734}$ & $\bm{80.4}$ & $\bm{76.5}$ & $\bm{0.722}$ & $\bm{79.6}$ & $\bm{75.2}$ & $\bm{0.710}$
			& \parbox{0pt}{\rule{0pt}{0ex+\baselineskip}} \\ [0ex]  	
			
			& \bf FT DeepSleepNet+& Yes & $\bm{80.2}$ & $\bm{76.6}$ & $\bm{0.722}$ & $\bm{80.1}$ & $\bm{76.6}$ & $\bm{0.721}$ & $\bm{81.5}$ & $\bm{77.5}$ & $\bm{0.738}$ & $\bm{77.4}$ & $\bm{74.1}$ & $\bm{0.682}$ & $\bm{76.0}$ & $\bm{71.4}$ & $\bm{0.661}$ & \parbox{0pt}{\rule{0pt}{0ex+\baselineskip}} \\ [0ex]  	
			
			& \it DT SeqSleepNet+ & Yes & $79.3$ & $73.0$ & $0.703$  & $73.1$ & $64.2$ & $0.615$ & $80.5$ & $75.6$ & $0.722$ & $67.2$ & $59.4$ & $0.531$ & $56.3$ & $48.4$ & $0.363$ & \parbox{0pt}{\rule{0pt}{0ex+\baselineskip}} \\ [0ex]  	
			
			& \it DT DeepSleepNet+ & Yes & $74.6$ & $67.4$ & $0.645$  & $71.6$ & $65.4$ & $0.600$ & $66.7$ & $61.3$ & $0.541$ & $70.0$ & $63.3$ & $0.586$ & $35.1$ & $31.0$ & $0.116$ & \parbox{0pt}{\rule{0pt}{0ex+\baselineskip}} \\ [0ex]  	
			
			& \it Scratch SeqSleepNet+ & No & $79.4$ & $74.5$ & $0.709$  & $79.6$ & $74.8$ & $0.711$ & $76.5$ & $70.6$ & $0.667$ & $78.6$ & $71.6$ & $0.693$ & $78.6$ & $71.6$ & $0.693$ &  \parbox{0pt}{\rule{0pt}{0ex+\baselineskip}} \\ [0ex]  	
			
			& \it Scratch DeepSleepNet+ & No & $73.8$ & $69.6$ & $0.634$ & $73.7$ & $67.6$ & $0.629$ & $72.4$ & $64.6$ & $0.603$ & $70.0$ & $65.9$ & $0.574$ & $70.0$ & $65.9$ & $0.574$ & \parbox{0pt}{\rule{0pt}{0ex+\baselineskip}} \\ [0ex]  	
			
			& SVM + Scattering Trans. \cite{Liu2019} & No & $-$ & $-$ & $-$ & $-$ & $-$ & $-$ & $78.6$ & $73.6$ & $0.695$ &  $-$ & $-$ & $-$ & $-$ & &  \parbox{0pt}{\rule{0pt}{0ex+\baselineskip}} \\ [0ex]  	
			
			& Decision trees. \cite{Sanders2014} & No & $-$ & $-$ & $-$ & $-$ & $-$ & $-$ & $75.0$ & $-$ & $-$ &  $-$ & $-$ & $-$ & $-$ & &  \parbox{0pt}{\rule{0pt}{0ex+\baselineskip}} \\ [0ex]  	
			
			\cline{1-18}
			
			\multirow{7}{*}{\begin{sideways}{\bf Surrey-cEEGrid}\end{sideways}} & \bf FT SeqSleepNet+ & Yes & $\bm{82.9}$ & $\bm{72.6}$ & $\bm{0.762}$ & $\bm{82.3}$ & $\bm{71.1}$ & $\bm{0.752}$ & $\bm{75.3}$ & $\bm{60.8}$ & $\bm{0.650}$ & $\bm{82.6}$ & $\bm{72.2}$ & $\bm{0.758}$ & $\bm{81.9}$ & $\bm{71.2}$ & $\bm{0.749}$
			& \parbox{0pt}{\rule{0pt}{0ex+\baselineskip}} \\ [0ex]  	
			
			& \bf FT DeepSleepNet+ & Yes & $\bm{71.1}$ & $\bm{59.7}$ & $\bm{0.588}$ & $\bm{77.8}$ & $\bm{66.5}$ & $\bm{0.687}$ & $\bm{58.2}$ & $\bm{42.8}$ & $\bm{0.391}$ & $\bm{77.5}$ & $\bm{66.6}$ & $\bm{0.682}$ & $\bm{81.7}$ & $\bm{70.5}$ & $\bm{0.745}$ & \parbox{0pt}{\rule{0pt}{0ex+\baselineskip}} \\ [0ex]  	
			
			& \it DT SeqSleepNet+  & Yes & $20.2$ & $14.7$ & $0.062$   & $19.4$ & $14.6$ & $0.051$ & $10.6$ & $9.1$ & $-0.015$ & $24.3$ & $20.5$ & $0.085$ & $24.1$ & $16.9$ & $0.090$ & \parbox{0pt}{\rule{0pt}{0ex+\baselineskip}} \\ [0ex]  	
			
			& \it DT DeepSleepNet+ & Yes & $38.4$ & $11.8$ & $0.025$  & $38.3$ & $11.7$ & $0.020$ & $38.4$ & $11.6$ & $0.012$ & $39.3$ & $25.4$ & $0.214$ & $38.9$ & $25.4$ & $0.195$
			& \parbox{0pt}{\rule{0pt}{0ex+\baselineskip}} \\ [0ex]  	
			
			& \it Scratch SeqSleepNet+ & No & $82.1$ & $67.6$ & $0.748$ & $81.5$ & $66.4$ & $0.739$ & $71.9$ & $55.2$ & $0.597$ & $81.3$ & $67.8$ & $0.737$ & $81.3$ & $67.8$ & $0.737$ &  \parbox{0pt}{\rule{0pt}{0ex+\baselineskip}} \\ [0ex]  	
			
			& \it Scratch DeepSleepNet+  & No & $65.6$ & $57.3$ & $0.535$ & $65.4$ & $57.4$ & $0.534$ & $42.5$ & $30.3$ & $0.195$ & $69.1$ & $60.0$ & $0.579$ & $69.1$ & $60.0$ & $0.579$  & \parbox{0pt}{\rule{0pt}{0ex+\baselineskip}} \\ [0ex]  	
			
			& Random Forests$^\diamond$ \cite{Mikkelsen2019} & No & $-$ & $-$ & $-$ & $72.0$ & $-$ & $0.600$ & $70.0$ & $-$ & $0.580$ & $-$ & $-$ & $-$ & $-$ & $-$ & $-$ & \parbox{0pt}{\rule{0pt}{0ex+\baselineskip}} \\ [0ex]  	
			\cline{1-18}
		\end{tabular}
	\end{center}
	\label{tab:performance_comaprison}
	\vspace{-0.3cm}
\end{table*}

\setlength\tabcolsep{1.5pt}
\begin{table*}[!t]
	\caption{Class-wise performance of the proposed transfer-learning systems and the baseline systems in terms of MF1.} 
	\tiny
	\vspace{-0.25cm}
	\begin{center}
		\begin{tabular}{|>{\arraybackslash}m{0.05in}|>{\arraybackslash}m{0.675in}|>{\centering \arraybackslash}m{0.225in}|>{\centering\arraybackslash}m{0.18in}|>{\centering\arraybackslash}m{0.18in}|>{\centering\arraybackslash}m{0.18in}|>{\centering\arraybackslash}m{0.18in}|>{\centering\arraybackslash}m{0.18in}|>{\centering\arraybackslash}m{0.18in}|>{\centering\arraybackslash}m{0.18in}|>{\centering\arraybackslash}m{0.18in}|>{\centering\arraybackslash}m{0.18in}|>{\centering\arraybackslash}m{0.18in}|>{\centering\arraybackslash}m{0.18in}|>{\centering\arraybackslash}m{0.18in}|>{\centering\arraybackslash}m{0.18in}|>{\centering\arraybackslash}m{0.18in}|>{\centering\arraybackslash}m{0.18in}|>{\centering\arraybackslash}m{0.18in}|>{\centering\arraybackslash}m{0.18in}|>{\centering\arraybackslash}m{0.18in}|>{\centering\arraybackslash}m{0.18in}|>{\centering\arraybackslash}m{0.18in}|>{\centering\arraybackslash}m{0.18in}|>{\centering\arraybackslash}m{0.18in}|>{\centering\arraybackslash}m{0.21in}|>{\centering\arraybackslash}m{0.21in}|>{\centering\arraybackslash}m{0.18in}|>{\centering\arraybackslash}m{0in} @{}m{0pt}@{}}
			\cline{2-28}
			\multicolumn{1}{c|}{} & \multirow{2}{*}{~~~~~~~~System}& \multirow{2}{*}{\makecell{Transfer\\learning}}&
			\multicolumn{5}{c|}{EEG$\cdot$EOG$\cdot$EMG$\mapsto$EEG$\cdot$EOG$\cdot$EMG} & \multicolumn{5}{c|}{EEG$\cdot$EOG$\mapsto$EEG$\cdot$EOG} & \multicolumn{5}{c|}{EEG$\mapsto$EEG} & \multicolumn{5}{c|}{EOG$\mapsto$EOG} & \multicolumn{5}{c|}{EEG$\mapsto$EOG} & \parbox{0pt}{\rule{0pt}{1ex+\baselineskip}} \\ [0ex]  	
			\cline{4-28}
			\multicolumn{1}{c|}{} &  & & W & N1 & N2 & N3 & REM & W & N1 & N2 & N3 & REM & W & N1 & N2 & N3 & REM & W & N1 & N2 & N3 & REM & W & N1 & N2 & N3 & REM &  \parbox{0pt}{\rule{0pt}{1ex+\baselineskip}} \\ [0ex]  	
			\cline{1-28}
			
			\multirow{6}{*}{\begin{sideways}{\bf Sleep-EDF-SC~~}\end{sideways}} & \bf FT SeqSleepNet+ & Yes &$-$ & $-$& $-$& $-$& $-$& $\bm{80.0}$ & $\bm{45.9}$ & $\bm{88.0}$ & $\bm{85.9}$ & $\bm{88.9}$& $\bm{85.4}$ & $\bm{50.9}$ & $\bm{88.8}$ & $\bm{86.4}$ & $\bm{86.5}$ & $\bm{75.1}$ & $\bm{46.4}$ & $\bm{86.3}$ & $\bm{80.3}$ & $\bm{87.3}$ & $\bm{72.8}$ & $\bm{40.3}$ & $\bm{84.9}$ & $\bm{78.7}$ & $\bm{84.8}$ & \parbox{0pt}{\rule{0pt}{0ex+\baselineskip}} \\ [0ex]  	
			
			& \bf FT DeepSleepNet+  & Yes &$-$ & $-$& $-$& $-$& $-$& $\bm{82.6}$ & $\bm{50.0}$ & $\bm{87.8}$ & $\bm{86.2}$ & $\bm{88.4}$ & 
			$\bm{81.0}$ & $\bm{50.5}$ & $\bm{88.2}$ & $\bm{86.9}$ & $\bm{87.2}$ & 
			$\bm{75.3}$ & $\bm{42.7}$ & $\bm{84.5}$ & $\bm{79.3}$ & $\bm{85.4}$ & 
			$\bm{75.7}$ & $\bm{41.9}$ & $\bm{83.9}$ & $\bm{78.1}$ & $\bm{84.5}$ & \parbox{0pt}{\rule{0pt}{0.25ex+\baselineskip}} \\ [0ex]  	
			
			& \it DT SeqSleepNet+ & Yes &$-$ & $-$& $-$& $-$& $-$& $63.2$ & $29.8$ & $84.9$ & $72.2$ & $60.2$& $74.1$ & $46.9$ & $86.9$ & $81.2$ & $83.8$ &  $67.7$ & $33.9$ & $79.3$ & $54.4$ & $60.4$& $51.4$ & $28.9$ & $61.0$ & $19.8$ & $51.6$ 
			& \parbox{0pt}{\rule{0pt}{0.25ex+\baselineskip}} \\ [0ex]  	
			
			& \it DT DeepSleepNet+ & Yes &$-$ & $-$& $-$& $-$& $-$&   $59.6$ & $30.6$ & $82.7$ & $80.5$ & $45.5$ & 
			$69.0$ & $31.9$ & $80.0$ & $74.6$ & $78.8$ & 
			$38.9$ & $15.1$ & $74.6$ & $78.8$ & $2.0$ & 
			$29.6$ & $11.5$ & $48.9$ & $75.1$ & $13.8$
			& \parbox{0pt}{\rule{0pt}{0.25ex+\baselineskip}} \\ [0ex]  	
			
			& \it Scratch SeqSleepNet+  & No &$-$ & $-$& $-$& $-$& $-$&   $75.0$ & $38.3$ & $86.8$ & $86.0$ & $85.0$ & $78.5$ & $37.1$ & $87.6$ & $86.2$ & $81.2$ &  $73.5$ & $25.8$ & $84.4$ & $77.7$ & $80.3$ & $73.5$ & $25.8$ & $84.4$ & $77.7$ & $80.3$ & \parbox{0pt}{\rule{0pt}{0.25ex+\baselineskip}} \\ [0ex]  	
			
			&  \it Scratch DeepSleepNet+ & No &$-$ & $-$& $-$& $-$& $-$&  $67.5$ & $47.9$ & $86.8$ & $86.8$ & $87.0$ & 
			$70.3$ & $48.1$ & $86.4$ & $84.6$ & $81.3$ & 
			$62.8$ & $33.1$ & $81.5$ & $74.8$ & $82.5$ & 
			$62.8$ & $33.1$ & $81.5$ & $74.8$ & $82.5$
			& \parbox{0pt}{\rule{0pt}{0.25ex+\baselineskip}} \\ [0ex]  	
			
			\cline{1-28}

			\multirow{7}{*}{\begin{sideways}{\bf Sleep-EDF-ST}\end{sideways}} & \bf FT SeqSleepNet+ & Yes & $\bm{80.5}$ & $\bm{54.0}$& $\bm{84.2}$ & $\bm{71.9}$ & $\bm{91.1}$ & $\bm{81.0}$ & $\bm{55.5}$ & $\bm{84.7}$ & $\bm{71.8}$ & $\bm{90.4}$ & $\bm{81.8}$ & $\bm{59.5}$ & $\bm{84.4}$ & $\bm{72.9}$ & $\bm{89.2}$ & $\bm{80.3}$ & $\bm{57.7}$ & $\bm{83.9}$ & $\bm{70.4}$ & $\bm{90.3}$ & $\bm{78.9}$ & $\bm{52.7}$ & $\bm{83.5}$ & $\bm{71.8}$ & $\bm{88.9}$ & \parbox{0pt}{\rule{0pt}{0.25ex+\baselineskip}} \\ [0ex]  	
			
			& \bf FT DeepSleepNet+& Yes & $\bm{80.8}$ & $\bm{55.7}$ & $\bm{82.9}$ & $\bm{71.3}$ & $\bm{87.9}$& $\bm{81.7}$ & $\bm{57.3}$ & $\bm{83.8}$ & $\bm{70.7}$ & $\bm{89.5}$ & $\bm{82.9}$ & $\bm{56.9}$ & $\bm{85.2}$ & $\bm{74.0}$ & $\bm{88.4}$ & $\bm{81.8}$ & $\bm{56.1}$ & $\bm{81.3}$ & $\bm{63.8}$ & $\bm{87.3}$ & $\bm{77.0}$ & $\bm{45.7}$ & $\bm{80.7}$ & $\bm{68.5}$ & $\bm{85.3}$ & \parbox{0pt}{\rule{0pt}{0.25ex+\baselineskip}} \\ [0ex]  	
			
			& \it DT SeqSleepNet+ & Yes &$76.9$ &$47.5$ & $85.4$& $68.5$ & $86.7$ &  $63.5$ & $39.6$ & $85.2$ & $65.5$ & $67.3$ & 
			$78.8$ & $56.0$ & $85.2$ & $69.2$ & $88.6$ &  
			$62.2$ & $44.0$ & $80.4$ & $61.9$ & $48.3$ & 
			$62.8$ & $41.1$ & $66.5$ & $17.7$ & $53.8$
			& \parbox{0pt}{\rule{0pt}{0.25ex+\baselineskip}} \\ [0ex]  	
			
			& \it DT DeepSleepNet+ & Yes &$67.9$ & $33.7$ & $81.9$ & $71.9$ &$81.6$& $73.2$ & $36.0$ & $78.4$ & $70.2$ & $69.3$ & 
			$66.8$ & $36.1$ & $73.6$ & $63.4$ & $66.6$ &  
			$62.4$ & $32.8$ & $79.6$ & $70.5$ & $71.2$ & 
			$32.4$ & $13.7$ & $40.6$ & $64.1$ & $4.3$
			& \parbox{0pt}{\rule{0pt}{0.25ex+\baselineskip}} \\ [0ex]  	
			
			& \it Scratch SeqSleepNet+ & No & $80.9$ & $49.2$ & $83.5$ & $70.0$ & $89.1$ & $82.0$ & $48.2$ & $83.3$ & $70.7$ & $89.6$ & 
			$80.3$ & $38.9$ & $82.0$ & $69.9$ & $81.6$ &
			$76.5$ & $38.4$ & $83.6$ & $72.2$ & $87.2$ & 
			$76.5$ & $38.4$ & $83.6$ & $72.2$ & $87.2$
			& \parbox{0pt}{\rule{0pt}{0.25ex+\baselineskip}} \\ [0ex]  	
			
			& \it Scratch DeepSleepNet+ & No & $72.0$ & $46.1$ & $78.3$ & $67.4$ &$84.4$ & $59.5$ & $47.4$ & $80.9$ & $72.3$ & $77.8$ & 
			$61.0$ & $40.3$ & $81.1$ & $67.1$ & $73.7$ & 
			$67.1$ & $45.7$ & $74.9$ & $64.3$ & $77.2$ & 
			$67.1$ & $45.7$ & $74.9$ & $64.3$ & $77.2$
			& \parbox{0pt}{\rule{0pt}{0.25ex+\baselineskip}} \\ [0ex]  	
			
			\cline{1-28}
			
			\multirow{7}{*}{\begin{sideways}{\bf Surrey-cEEGrid}\end{sideways}} & \bf FT SeqSleepNet+ & Yes & $\bm{91.6}$ & $\bm{81.3}$ & $\bm{27.2}$ & $\bm{81.3}$ & $\bm{81.3}$ & $\bm{90.9}$ & $\bm{79.9}$ & $\bm{23.1}$ & $\bm{81.4}$ & $\bm{80.2}$ & $\bm{90.6}$ & $\bm{58.0}$ & $\bm{10.6}$ & $\bm{71.5}$ & $\bm{73.4}$ & $\bm{91.2}$ & $\bm{78.8}$ & $\bm{26.6}$ & $\bm{81.1}$ & $\bm{83.3}$ & $\bm{91.3}$ & $\bm{76.0}$ & $\bm{26.4}$ & $\bm{81.0}$ & $\bm{81.0}$ & \parbox{0pt}{\rule{0pt}{0.25ex+\baselineskip}} \\ [0ex]  	
			
			& \bf FT DeepSleepNet+ & Yes & $\bm{80.4}$ & $\bm{63.3}$ & $\bm{10.3}$ & $\bm{67.5}$ & $\bm{77.1}$ & $\bm{86.7}$ & $\bm{70.9}$ & $\bm{15.5}$ & $\bm{77.1}$ & $\bm{82.5}$ & $\bm{74.9}$ & $\bm{23.2}$ & $\bm{5.7}$ & $\bm{47.0}$ & $\bm{63.4}$ & $\bm{87.0}$ & $\bm{66.6}$ & $\bm{22.0}$ & $\bm{77.7}$ & $\bm{79.8}$ & $\bm{90.2}$ & $\bm{78.5}$ & $\bm{20.2}$ & $\bm{80.9}$ & $\bm{82.8}$ & \parbox{0pt}{\rule{0pt}{0.25ex+\baselineskip}} \\ [0ex]  	
			
			& \it DT SeqSleepNet+  & Yes & $57.0$ & $1.6$ & $11.9$ & $3.1$ & $0.0$ & $54.4$ & $0.4$ & $12.1$ & $6.0$ & $0.0$ & $29.2$ & $0.6$ & $9.9$ & $3.3$ & $2.5$ &  $46.7$ & $12.1$ & $13.7$ & $30.1$ & $0.0$ &  $67.9$ & $2.8$ & $10.1$ & $3.7$ & $0.1$ & \parbox{0pt}{\rule{0pt}{0.25ex+\baselineskip}} \\ [0ex]  	
			
			& \it DT DeepSleepNet+ & Yes & $57.6$ & $0.0$ & $1.3$ & $0.0$ & $0.0$ & $57.0$ & $0.0$ & $1.7$ & $0.0$ & $0.0$ & $57.1$ & $1.2$ & $0.0$ & $0.0$ & $0.0$ & $67.0$ & $0.4$ & $17.2$ & $42.4$ & $0.0$ & $66.2$ & $3.1$ & $14.6$ & $43.3$ & $0.0$ & \parbox{0pt}{\rule{0pt}{0.25ex+\baselineskip}} \\ [0ex]  	
			
			& \it Scratch SeqSleepNet+ & No & $90.8$ & $81.1$ & $5.7$ & $80.2$ & $80.0$ & $90.4$ & $79.9$ & $3.3$ & $79.9$ & $78.7$ & $88.2$ & $50.5$ & $1.0$ & $67.4$ & $68.8$ &  $90.3$ & $79.6$ & $12.3$ & $79.5$ & $77.5$ & $90.3$ & $79.6$ & $12.3$ & $79.5$ & $77.5$ & \parbox{0pt}{\rule{0pt}{0.25ex+\baselineskip}} \\ [0ex]  	
			
			& \it Scratch DeepSleepNet+  & No & $75.0$ & $61.4$ & $23.4$ & $64.4$ & $62.4$ & $72.9$ & $59.4$ & $24.0$ & $68.7$ & $61.8$ & $57.6$ & $12.3$ & $6.9$ & $23.1$ & $51.4$ & $75.2$ & $66.1$ & $19.7$ & $73.0$ & $66.0$ & $75.2$ & $66.1$ & $19.7$ & $73.0$ & $66.0$ & \parbox{0pt}{\rule{0pt}{0.25ex+\baselineskip}} \\ [0ex]  	
			\cline{1-28}
		\end{tabular}
	\end{center}
	\label{tab:performance_classwise}
	\vspace{-0.4cm}
\end{table*}

\subsubsection{Performance comparison on the target domains}

To justify the necessity of transfer learning, in Table \ref{tab:performance_comaprison} we compare the finetuning overall performance against those of the scratch models and \emph{direct transfer} (i.e. applying the pretrained models in the target domains without finetuning) across the target domains and the transfer learning scenarios. In addition, the obtained results are also contrasted to those reported in previous works to quantify the efficiency of the proposed transfer learning approach. As the transfer learning results vary depending on the finetuning strategies, for simplicity, out of different finetuning strategies, we retained the SPB+softmax one as the representative for comparison given its consistent finetuning results (see Figures  \ref{fig:seqsleepnet_finetune} and \ref{fig:deepsleepnet_finetune}).  In practice, the finetuning strategies could be viewed as a hyper-parameter and determined via cross-validation. We should bring to readers' attention a large body of works, such as \cite{Biswal2018a,Stephansen2018,Sun2017,Olesen2018}, that yielded an accuracy level on (extremely) large databases similar to that of our proposed systems. However, comparison to these results is not the main focus of this work; furthermore, such a comparison would be incompatible and, hence, does not offer much meaning.

Between SeqSleepNet+ and DeepSleepNet+, the former outperforms the latter in most of the cases in Table \ref{tab:performance_comaprison}. With scratch training, SeqSleepNet+ results in an average accuracy gain of $1.7\%$, $6.6\%$, and $17.3\%$ over DeepSleepNet+ on Sleep-EDF-SC, Sleep-EDF-ST, and Surrey-cEEGrid, respectively. This is consistent with the findings from the source domain (i.e. the MASS database) in Table \ref{tab:performance_source} and in \cite{Phan2019a}. With transfer learning, SeqSleepNet+ also obtains better performance than DeepSleepNet+ with, improving the overall accuracy by $0.8\%$, $1.5\%$, and $7.7\%$ on Sleep-EDF-SC, Sleep-EDF-ST, and Surrey-cEEGrid, respectively. These results suggest that DeepSleepNet+ is harder to train and finetune than SeqSleepNet+, especially when the data is small, partly due to its large model footprint \cite{Supratak2017} and partly due to its reliance on raw signal inputs. However, the results in Table \ref{tab:performance_comaprison} show significant gains obtained by both the finetuned models over their scratch counterparts. On the one hand, averaging over all transfer learning scenarios, finetuning SeqSleepNet+ leads to an absolute accuracy gain of $2.5\%$, $2.0\%$, and $1.4\%$ on Sleep-EDF-SC, Sleep-EDF-ST, and Surrey-cEEGrid, respectively. Those gains of DeepSleepNet+ are even larger, reaching $3.4\%$, $7.1\%$, and $10.9\%$, respectively, mainly because of the poor performance of the scratch DeepSleepNet+ on Sleep-EDF-ST and Surrey-cEEGrid. Interestingly, transfer learning helps compensate for the lack of training data, evidenced by the observation that the accuracy on Sleep-EDF-SC achieved by the finetuned SeqSleepNet+ is on par with that of MASS (cf. Table \ref{tab:performance_source}) even though the number of subjects is ten times smaller. On the other hand, despite the heavy data mismatch in the cross-domain scenario, transferring the information of EEG data in the source domain to EOG data in the target domains still yields significant accuracy gains: $1.0\%$ and $7.4\%$ on average with SeqSleepNet+ and DeepSleepNet+, respectively. Interestingly, with the accuracy consistently around $80\%$ obtained from the secondary EOG via DeepSleepNet+'s transfer learning, it is promising to be used as an alternative for EEG in single-channel sleep staging.

Directly applying the pretrained models in the target domains without finetuning results in suboptimal performance in many cases. Averaging over the same-modality transfer learning scenarios, the pretrained SeqSleepNet+ model with direct transfer obtains an accuracy with $10.3\%$, $5.7\%$, and $62.2\%$ lower than those obtained by the finetuned models on Sleep-EDF-SC, Sleep-EDF-ST, and Surrey-cEEGrid, respectively. Those gaps in case of DeepSleepNet+ are $16.8\%$, $9.1\%$, and $32.6\%$, respectively. The direct transfer's results are particularly poor under heavy data mismatch conditions, such as the EEG$\mapsto$EOG scenario and the EEG$\mapsto$EEG scenario in Surrey-cEEGrid. It is reasonable as substantial differences in characteristics of the source domain and the target domain cause discrepancy in the feature-learning parts of the pretrained models in the target domain. As a consequence, finetuning is essential. Similar findings are also reflected in the class-wise performance (in terms of MF1) in Table \ref{tab:performance_classwise}.

The proposed transfer learning approach also outperforms all previous works and set state-of-the-art performance on all three target databases. On Sleep-EDF-SC, with the accuracies of $84.3\%$ (two-channel EEG$\cdot$EOG) and $85.2\%$ (single-channel EEG) obtained by the transfer learning based SeqSleepNet+, the system yields absolute accuracy gains of $2.0\%$ and $3.2\%$ over the best non-transfer-learning systems, Multitask 1-max CNN \cite{Phan2019b} ($82.3\%$) and DeepSleepNet \cite{Supratak2017} ($82.0\%$), respectively. Those respective gains achieved by the transfer learning based DeepSleepNet+ are $2.3\%$ and $2.4\%$. Large margins, $7.5\%$ and $7.8\%$, are seen when contrasting the proposed SeqSleepNet+ and DeepSleepNet+ systems with the existing transfer learning approach based on ResNet \cite{Andreotti2018} and VGGNet \cite{Vilamala2017}. These results suggest that the quality of the base model plays an important role in transfer learning for sleep staging. The results obtained by the proposed systems are also better than the personalization results in \cite{Mikkelsen2018} even though cohort transfer learning here is more challenging than personalized transfer learning as, with the former, we do not have access to test subjects' data during training. Similar to Sleep-EDF-SC, both proposed systems are superior to previous works on Sleep-EDF-ST. However, on Surrey-cEEGrid, while the transfer learning based SeqSleepNet+ uplifts the accuracy by a margin of $10.3\%$ in two-channel EEG$\cdot$EOG and $5.3\%$ in single-channel EEG compared to the seminal work in \cite{Mikkelsen2019}, the DeepSleepNet+ experiences an accuracy drop of $11.8\%$ in single-channel EEG even though $5.8\%$ absolute accuracy gain is seen in two-channel EEG$\cdot$EOG.

\subsubsection{Influence of the number of finetuning subjects}

This section investigates the influence of the amount of the target-domain data to the network finetuning. Considering the EEG$\cdot$EOG$\mapsto$EEG$\cdot$EOG scenario and the entire-network finetuning strategies for this investigation. For a target domain, we randomly selected 25\% of the subjects as the test subjects while the remaining subjects were used for finetuning. A pretrained network was finetuned using data from the finetuning set of $N$ subjects for 500 finetuning steps and the test accuracy was recorded during the finetuning process. Starting with the finetuning set of $N=1$ subject, we repeated this procedure and added two more subjects into it at each iteration.

\begin{figure} [!t]
	\centering
	\includegraphics[width=0.9\linewidth]{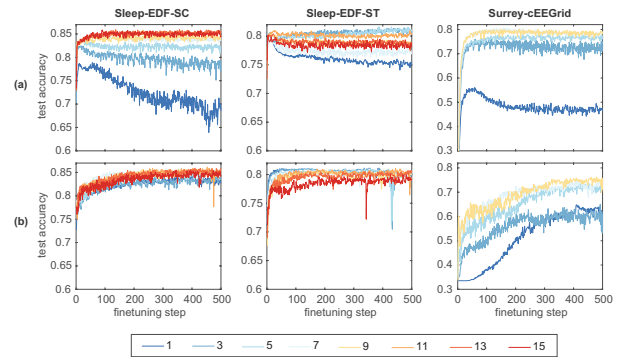}
	\caption{The learning curves obtained from the test data with varying the number of finetuning subjects. (a) SeqSleepNet+ and (b) DeepSleepNet+.}
	\label{fig:influence_nosub}
	\vspace{-0.25cm}
\end{figure}

Figure \ref{fig:influence_nosub}  shows the learning curves recorded with varying number of finetuning subjects. The learning curves present a strong impact of the number of finetuning subjects on SeqSleepNet+ while such influence on DeepSleepNet+ is less noticeable, except for Surrey-cEEGrid. It is rational if these results are linked to the networks' finetuning behaviors. While a pretrained SeqSleepNet+ requires its feature-learning parts to be adapted into the target domains, this requirement is not mandatory for DeepSleepNet+, except for the cEEGrid data (see Section \ref{sssec:transfer_learning_effects}). And when the feature-learning parts need to be adjusted, less finetuning data make the networks converge to more subject-specific solutions, i.e. overfitting. On the contrary, more finetuning data allow the feature learning parts to converge to more generalizable solutions. This is supported by the SeqSleepNet+'s learning curves on the Sleep-EDF-SC and Surrey-cEEGrid domain, and DeepSleepNet+'s learning curves on the Surrey-cEEGrid domain. From these curves, we also speculate that when the feature-learning parts of a network needs to be adapted to a target domain, a generalizable solution can be obtained with the number of finetuning subjects being around 11-13. Particularly, the learning curves on Sleep-EDF-ST appears to be counter-intuitive as more finetuning subjects occasionally result in lowering learning curves. These irregularities can be explained by the fact that the Sleep-EDF-ST population has a very wide range of age, 18-79. As sleep patterns change with age \cite{Skeldon2016}, depending the age range of the test subjects, including a subject whose age is far from that range would hurt more than help. Further studies how to determine and select candidates from a population that are most beneficial for a finetuning task.

\vspace{-0.25cm}
\subsection{Discussion}
It is worth mentioning that, although we focused on studying with small cohorts in this work, the presented transfer learning approach would also be useful for a sleep study with a larger cohort. On the one hand, it only requires the data of a handful of subjects to be labelled, avoiding the burden of manual scoring the entire cohort. On the other hand, finetuning a pretrained model is generally much faster than training a model from scratch, as illustrated in Figure \ref{fig:learning_curve}. This is because the pretrained model has reached already a reasonable accuracy. As a result, it is able to converge after a few additional finetuning epochs. On the downside, it is worth noting that still data from a number of subjects is needed for the validation purpose and future works should explore regularization methods, such as Kullback–Leibler divergence \cite{Phan2020a}, to eliminate this requirement.

\vspace{-0.25cm}
\section{Conclusion}
We presented a deep transfer learning approach to address the problem of insufficient data in many sleep studies and to improve automatic sleep staging performance on small cohorts. The SeqSleepNet+ and DeepSleepNet+ derived from the presented generic sequence-to-sequence sleep staging framework were employed to surpass data mismatch and enable transferring information from the source domain to the target domain. The networks were trained in the source domain and then finetuned in the target domains to complete knowledge transfer. Experiments were conducted with different finetuning strategies, transfer learning scenarios, and target domains. The experimental results showed that via transfer learning, the sleep staging performance was significantly improved across all learning cases over the scratch models trained solely on the target domains. The results also revealed the different behaviors of two SeqSleepNet+ and DeepSleepNet+ models in transfer learning. The former was found more consistent and stable and outperformed the latter in most of the transfer learning experiments. The number of subjects required for finetuning also varied between the two networks, however, overall, a small number of finetuning subjects was needed for the networks to converge to a generalizable solution.

\vspace{-0.25cm}
\section*{Acknowledgment}
\vspace{-0.1cm}
This research received funding from the Flemish Government (AI Research Program). Maarten De Vos is affiliated to Leuven.AI - KU Leuven institute for AI, B-3000, Leuven, Belgium. We gratefully acknowledge the support of NVIDIA Corporation with the donation of the Titan V GPU used for this research. We would like to thank Dr. Kaare Mikkelsen for sharing the Surrey-cEEGrid database.

\vspace{-0.35cm}
\bibliographystyle{IEEEbib}
\bibliography{bibliography}

\end{document}